# Urban functional knowledge transfer through global representations amid uneven data

Junyuan Liu[a,#], Xinglei Wang[a,#], Guangsheng Dong[b,a,*], Hao Sha[b], Zichao Zeng[a], Zhicheng Deng[a], Jiazhuang Feng[a], Xiang Wang[b], Tao Cheng[a,*]

[a] SpaceTimeLab, Department of Civil, Environmental and Geomatic Engineering, University College London, London, WC1E 6BT, United Kingdom

[b] State Key Laboratory of Information Engineering in Surveying, Mapping and Remote Sensing, Wuhan University, Wuhan, 430079, China

[#] These authors contributed equally.

[*] Correspondence to Guangsheng Dong & Tao Cheng (guangshengdong@whu.edu.cn; tao.cheng@ucl.ac.uk).

## Abstract

Urban functional observations are unevenly distributed worldwide, limiting analysis of how places are organised and used. AlphaEarth Foundations (AEF) provides a globally consistent physical representation. Whether this representation can support transfer of functional knowledge to locations with sparse or absent functional observations, and what governs transfer, remain unclear. We introduce POI-aligned Urban Embedding (PURE), which learns physical–functional relationships between POI-derived semantics and AEF. Across five urban analysis tasks covering 142 countries and territories, PURE increases mean $R^2$ from 0.792 to 0.830, with gains across widely different levels of POI availability. When target-city POIs are excluded, country-level alignment retains 87.9% of its performance gain. Cross-country experiments show that transfer is selective rather than universal, with larger gains between physically similar and geographically closer contexts, particularly among pairs within the same continent. Our findings reveal spatially structured urban functional knowledge transfer, offering a route to reduce reliance on dense local observations.

## Introduction

Urban research is shaped by an uneven geography of data availability. Existing studies are concentrated in a limited number of countries and major metropolitan areas, while cities in the Global South, smaller urban centres and rapidly growing settlements are underrepresented[1–3]. These knowledge gaps partly reflect disparities in the underlying data infrastructure. The completeness of global building data varies with geography, city size and human-development conditions[4]. Estimates of urban land differ across definitions and spatial scales[5], and crowdsourced observations reflect uneven social and geographic participation[6]. Although harmonised global databases have expanded the scope of comparative urban research, spatially explicit and multi-thematic information remains incomplete and fragmented across much of the global urban system[7]. These limitations are particularly consequential for places that are difficult to observe through conventional data systems, including informal settlements, where fine-scale mapping reveals substantial inequalities that aggregate urban statistics often obscure[8]. Functional observations describing how places are organised and used are similarly uneven, particularly when they depend on locally collected or contributed data. Unequal availability of functional observations can translate into unequal capacity to represent, compare and analyse urban function worldwide[9].

Earth observation provides a globally available source of information that is less dependent on local data collection[10]. Satellite imagery and aerial observations have been widely used to estimate poverty, wealth, consumption and economic activity at spatial resolutions difficult to achieve through conventional surveys alone[11–19]. Relationships learned from imagery in one set of locations can also remain informative elsewhere[15], suggesting that globally available observations can support knowledge transfer to places with limited reference data. More recently, reusable Earth representations have shifted this approach from task-specific models towards shared spatial encodings that can support multiple socioeconomic and environmental applications[17,20–22]. Earth foundation models extend this idea by learning general representations from large and diverse Earth observation datasets, providing rich and consistent descriptions of Earth-surface conditions across regions and tasks. AlphaEarth Foundations (AEF), for example, provides a globally consistent representation of annual land-surface conditions[20]. Such representations therefore provide a common physical description of environments across otherwise uneven data geographies.

A globally consistent physical representation, however, is not a complete urban representation. Earth observation provides information about land cover, built intensity, vegetation, roads and spatial structure, but these observable characteristics do not fully describe how urban places are organised and used[23–26]. Areas with similar physical form may support residential, commercial, industrial, transport, educational or medical functions, resulting in different patterns of population concentration, economic activity and environmental conditions[27–32]. Recent studies combining remote sensing with points of interest (POIs) or other social-sensing data have obtained more complete representations of urban land use and functional organisation than either source can provide alone[23,27–29]. Global Earth representations and functional observations therefore describe complementary dimensions of urban environments. The former provides broad and consistent information about observable physical conditions, whereas the latter make the organisation and use of places more explicit.

POIs are widely used to characterise the functions, amenities and activities associated with urban places[33–35]. But their availability and quality vary substantially across data sources and locations[33,34,36,37]. POI and related crowdsourced urban databases differ in geographic coverage, completeness, category systems, attribute detail and the populations and activities represented by their contributing platforms[33,36,37]. These differences can affect global comparisons of service availability and accessibility, while location-based observations may leave important dimensions of urban experience unobserved[34,35,37]. Enriching globally available physical representations with local functional observations may still reproduce existing inequalities in urban data provision if every city requires a comprehensive local POI database. This limitation is particularly important because physical observations alone cannot fully distinguish the uses and socioeconomic functions of urban infrastructure[38,39]. Cross-city learning and global Earth representations offer a potential means of transferring information from locations with dense functional observations to those where such observations are sparse[40,41]. However, it remains unclear whether physical–functional relationships learned through POI supervision in one geographic context remain informative where local functional observations are sparse or absent, and whether the strength of such transfer varies systematically with source–target context.

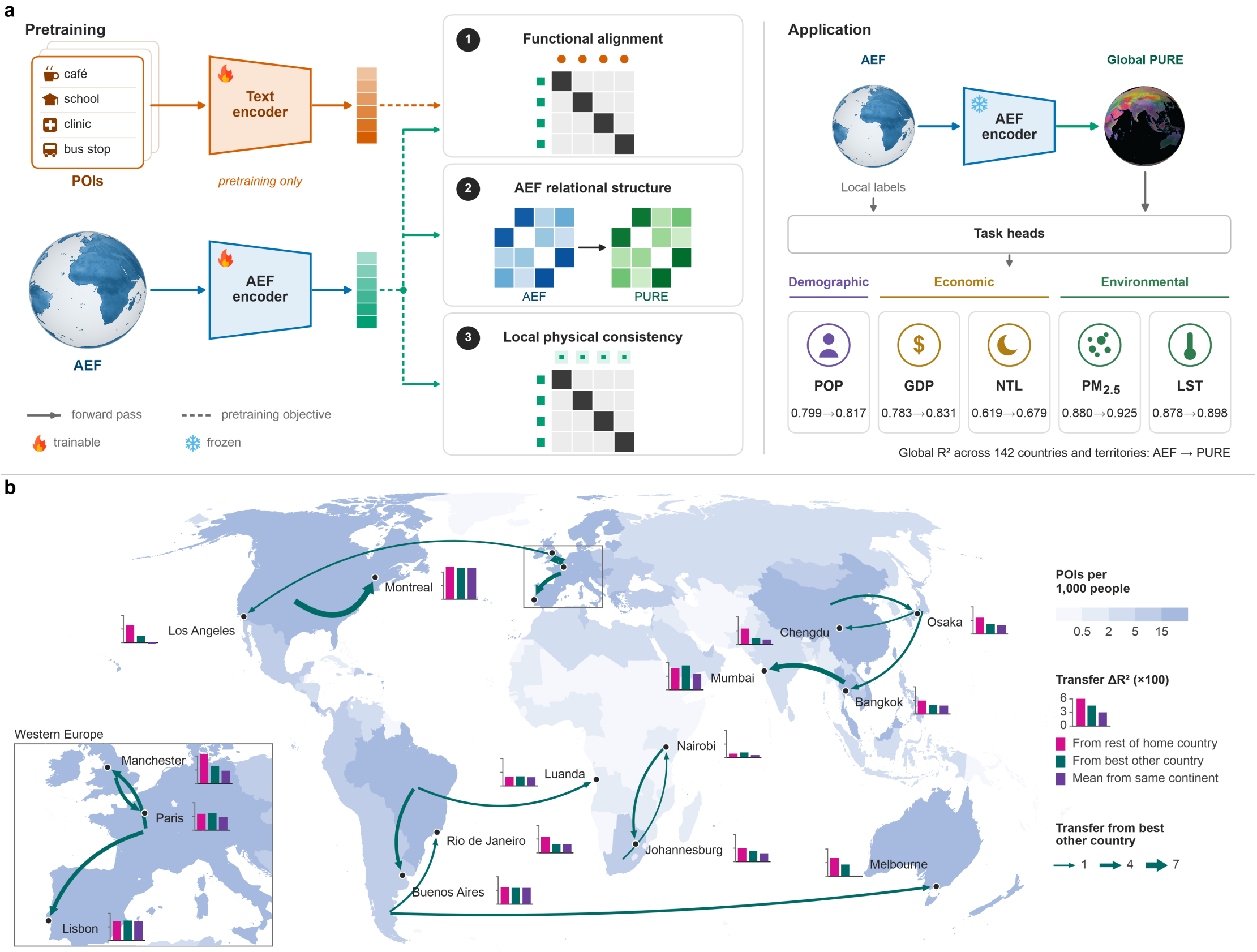

**Fig. 1. PURE framework and the transfer of urban functional knowledge across uneven data geographies. a,** PURE pretraining and application. During pretraining, a text encoder embeds POI descriptions, and an AEF encoder maps AEF features to PURE embeddings. Both encoders are trainable. Three objectives shape the AEF encoder: (1) functional alignment, which matches each cell with its POI description; (2) AEF relational structure, which preserves the pairwise relations of the frozen AEF features; and (3) local physical consistency, which matches each cell with its 3 × 3 neighbourhood. The text encoder is used only during pretraining. In application, the frozen AEF encoder generates global PURE embeddings from AEF features and geographic position only, and task-specific heads are trained with local labels. Numbers give the global $R^2$ of AEF and PURE for each task, the mean of country-level $R^2$ across 142 countries and territories. **b,** Transfer of functional knowledge to 15 target cities. Countries are shaded by POIs per 1,000 people in the PURE training data. Bars beside each city give the gain in five-task mean $R^2$ over AEF (×100) from training PURE on POIs from the rest of the city's home country, with all target-city POIs withheld (Country LOCO); from its best-performing other source country; and, averaged, from the other source countries on the same continent. All bars share one height scale. Arrows run from the interior of the best-performing other source country to the city, with width proportional to its gain (Fig. 4c). The inset enlarges Western Europe.

Here we show that globally available Earth representations can support the transfer of urban functional knowledge across uneven data geographies, and that this transfer is spatially structured. We introduce POI-aligned Urban Embedding (PURE), which aligns POI-derived functional semantics with AEF (Fig. 1a). The model is pretrained globally with over 84 million POIs. Across five tasks predicting demographic (population, POP), economic (gross domestic product, GDP), activity (nighttime lights, NTL) and human-environmental ($PM_{2.5}$ and land surface temperature, LST) outcomes in 142 countries and territories, we assess whether functional alignment adds predictive information beyond AEF and whether these gains depend on local POI availability. We then withhold target-city POIs and evaluate representation-level transfer across fifteen source countries and fifteen target cities spanning six continents (Fig. 1b). Functional alignment improves performance across widely varying levels of POI availability, and most of the country-level gain persists even without target-city POIs. Cross-country transfer is selective, with stronger gains between physically similar and geographically closer contexts. We further evaluate PURE for small-area socioeconomic prediction and metropolitan commuting flow generation and release the resulting global embeddings as an open resource for broader urban studies.

# Results

## 1. Functional information enriches a globally consistent physical basis for urban analysis

During pretraining of PURE (Fig. 1a), an AEF encoder learned to match grid cells with text descriptions of their POIs. Two further objectives preserved the relational structure of the original AEF features and the consistency of each cell with its immediate surroundings. After pretraining, the frozen encoder generated PURE embeddings from AEF features and geographic position alone, so that POIs were not required where the representation was applied. For each country and task, a prediction head was then trained on these embeddings with local target observations, and an identical head was trained on AEF with the same observations, spatially blocked data partitions and five random seeds. Performance was measured with the coefficient of determination ($R^2$), and the global $R^2$ of each task was calculated as the mean of country-level values.

Adding POI-derived functional semantics improved predictive performance across all five tasks, increasing mean $R^2$ from 0.792 to 0.830 and reducing unexplained variance by 18.2% overall (Fig. 1a). The reduction in unexplained variance was largest for $PM_{2.5}$ (37.4%) and GDP (21.8%), followed by LST (16.3%), NTL (15.7%) and POP (8.8%). Both auxiliary pretraining objectives contributed to this gain, as the five-task mean $R^2$ fell to 0.821 without AEF relational structure and to 0.809 without local physical consistency (Supplementary Table 8). The advantage did not depend on the evaluation metric or prediction head, as PURE also reduced standardised mean absolute error for every task and improved $R^2$ with linear heads (Supplementary Tables 4 and 7). Gains were also evident at the country level (Fig. 2a). Median $R^2$ increased for every task, and PURE improved $R^2$ in 678 of the 710 country–task combinations, with 23 of the 32 declines smaller than 0.01. The consistency of these gains across demographic, economic, activity and human-environmental outcomes indicates that functional alignment contributed information not fully recoverable from the physical representation alone. POIs can, for example, distinguish shops, services and workplaces from physically similar residential buildings. Rather than replacing the information encoded by physical form, POI-derived semantics enriched the representation with information about how urban places are organised and used. Gains also tended to be larger where AEF performance was weaker (Spearman's $\rho = -0.55$, $P < 0.001$), suggesting that this functional information was most valuable where physical form alone was least informative.

Country-level gains in $R^2$ ($\Delta R^2$ = PURE − AEF) were geographically widespread (Fig. 2b–f). All 142 countries and territories had a positive five-task mean $\Delta R^2$, and 112 improved in all five tasks. PURE improved $R^2$ in 138 countries and territories for GDP, 129 for POP, 131 for NTL, 141 for $PM_{2.5}$ and 139 for LST, and in at least 90% of the country–task combinations on each continent. For example, GDP $R^2$ increased from 0.593 to 0.656 in China and from 0.837 to 0.875 in the United States. The size of the gains differed among tasks and regions. GDP gains were substantial on every continent, with continental mean $\Delta R^2$ of 0.033–0.056, whereas POP gains were smaller (0.008–0.023). Gains for $PM_{2.5}$ and LST were largest in Europe (mean $\Delta R^2$ of 0.099 and 0.050, respectively), where AEF performance for these outcomes was weaker than on other continents (Supplementary Table 3). Declines were few and scattered across regions. The benefits of functional alignment were therefore broadly distributed across geographic contexts and urban outcomes.

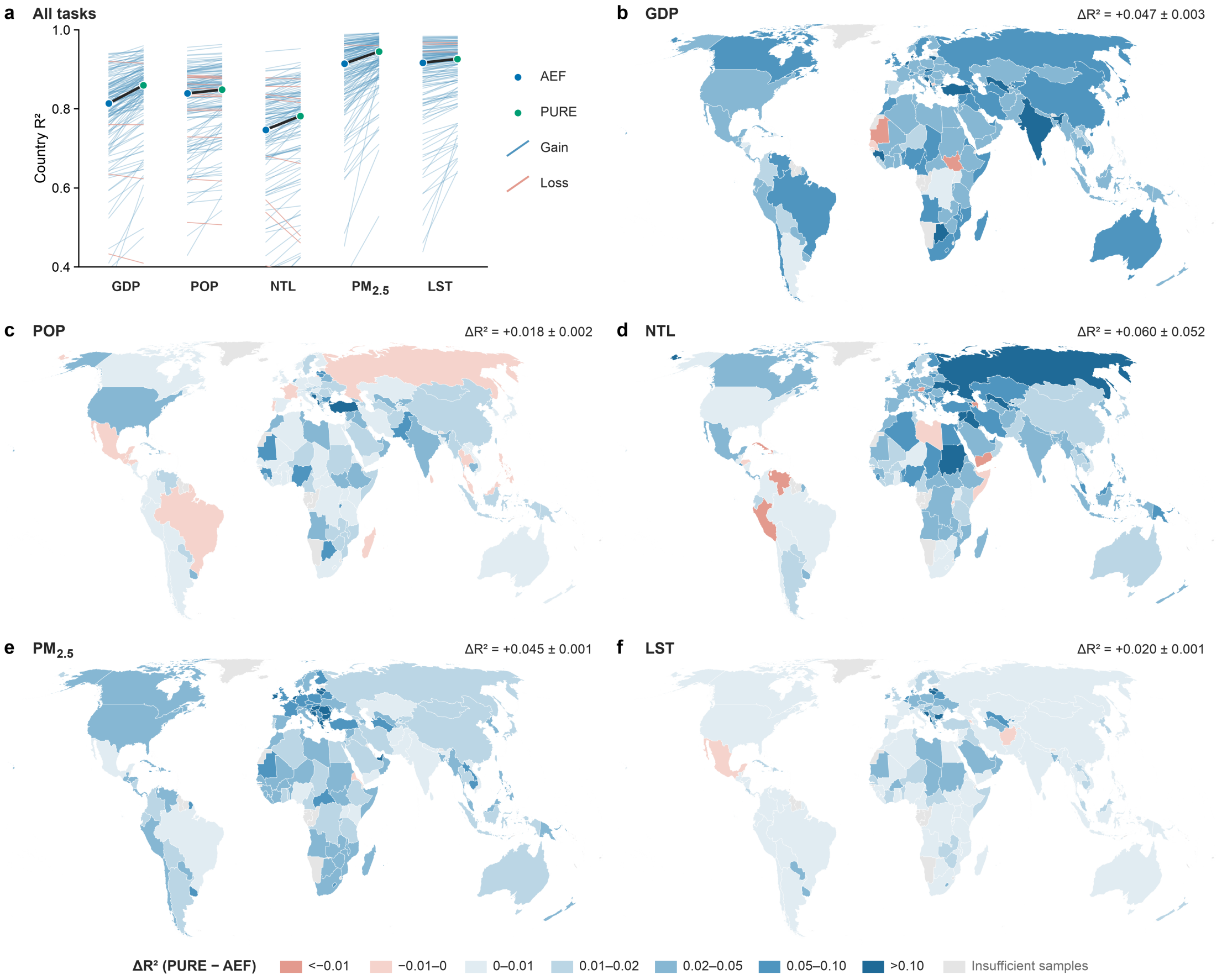

**Fig. 2. Country-level performance of AEF and PURE across five urban tasks. a,** Country-level $R^2$ of AEF and PURE for each task. Each thin line connects the AEF (left) and PURE (right) $R^2$ of one country or territory; blue lines indicate higher $R^2$ with PURE and red lines lower $R^2$. Bold segments connect the median $R^2$ of AEF (blue) and PURE (green). Lines below $R^2 = 0.4$ are clipped (29 of 710 country–task pairs); medians use all values. **b–f,** Country-level gain in $R^2$ from PURE over AEF ($\Delta R^2$ = PURE − AEF) for GDP (**b**), POP (**c**), NTL (**d**), $PM_{2.5}$ (**e**) and LST (**f**). Values above each map give the cross-country mean $\Delta R^2$ (mean ± s.d. across five random seeds). Grey indicates countries and territories outside the evaluation cohort, which required more than 10,000 valid settlement cells for POP and more than 1,000 for each other task (see 'Dependence on POI availability' in Methods). Country-level $R^2$ values are averaged across five random seeds; $n = 142$ countries and territories per task.

## 2. Functional-alignment gains extend across uneven POI geographies

PURE reduced cross-country variability in predictive performance for every task, measured as the median absolute deviation (MAD) of country-level $R^2$ (Fig. 3a). The reduction was largest for $PM_{2.5}$ (36.6%) and GDP (18.4%), followed by LST (12.7%), POP (4.7%) and NTL (2.4%), indicating that functional alignment narrowed performance differences among countries as well as raising average performance. These improvements occurred despite substantial geographic unevenness in the functional observations available for representation learning.

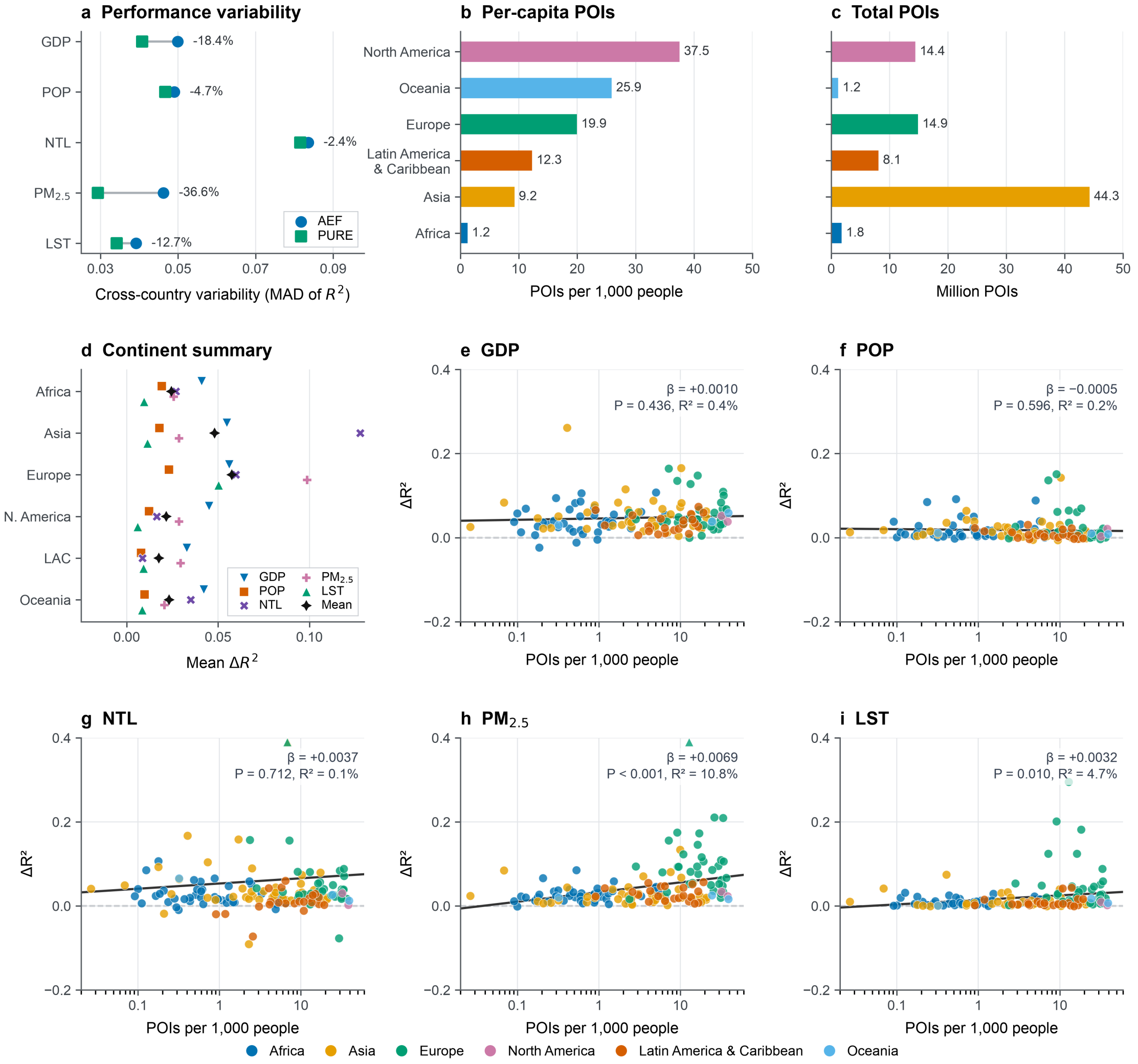

**Fig. 3. POI availability and the distribution of PURE gains. a,** Cross-country variability in predictive performance, measured as the median absolute deviation (MAD) of country-level $R^2$ for AEF (blue circles) and PURE (green squares); percentages give the change with PURE. **b–c,** POIs per 1,000 people (**b**) and total POI counts (**c**) by continent. **d,** Mean country-level $\Delta R^2$ by continent and task; black stars indicate the five-task mean. LAC, Latin America and the Caribbean. **e–i,** Country-level $\Delta R^2$ against POIs per 1,000 people (logarithmic axis) for GDP (**e**), POP (**f**), NTL (**g**), $PM_{2.5}$ (**h**) and LST (**i**), coloured by continent. Lines show log-linear ordinary least-squares fits; $\beta$ is the change in $\Delta R^2$ per doubling of POI density, reported with its two-sided $P$ value and the variance explained ($R^2$). Triangles indicate values outside the displayed y-axis range; fits use all original values. $n$ = 142 countries and territories per task.

POI density varied approximately 31-fold across continents, from 1.2 POIs per 1,000 people in Africa to 37.5 in North America. Europe and Oceania were also substantially more densely covered than Asia and Latin America and the Caribbean (Fig. 3b). Absolute POI counts followed a different pattern because of differences in population size (Fig. 3c). Asia contained the largest number of POIs, at 44.3 million, despite its comparatively modest POI density.

Despite this uneven observational basis, gains from functional alignment were distributed across all continents and tasks (Fig. 3d). Country-level analyses showed that the magnitude of these gains was not consistently associated with local POI density (Fig. 3e–i). The relationship between $\Delta R^2$ and log-transformed POIs per 1,000 people, expressed as the change in $\Delta R^2$ per doubling of POI density ($\beta$), was not statistically significant for GDP ($\beta = 0.0010$, $P = 0.436$), POP ($\beta = -0.0005$, $P = 0.596$) or NTL ($\beta = 0.0037$, $P = 0.712$). $PM_{2.5}$ ($\beta = 0.0069$, $P < 0.001$) and LST ($\beta = 0.0032$, $P = 0.010$) showed positive pooled associations. However, the corresponding models explained only 10.8% and 4.7% of the cross-country variation in PURE gain, respectively. POI density therefore had limited and task-dependent explanatory power rather than a consistent association across all five outcomes.

Positive gains occurred in both POI-sparse and POI-rich settings. Ethiopia, Uzbekistan and Papua New Guinea had only 0.091, 0.406 and 0.320 POIs per 1,000 people, respectively, yet achieved five-task mean $\Delta R^2$ values of 0.022, 0.117 and 0.025. Positive mean gains were also observed in POI-rich countries, including the United Kingdom, with 36.5 POIs per 1,000 people and a mean $\Delta R^2$ of 0.034, and Germany, with 24.7 POIs per 1,000 people and a mean $\Delta R^2$ of 0.044. Functional alignment therefore remained beneficial across more than two orders of magnitude of country-level POI density.

These results show that the benefits of functional alignment were geographically widespread rather than confined to countries with dense functional observations. Detailed POI-corpus coverage, global target support and continent-level estimates are reported in Supplementary Tables 1–3.

## 3. Urban functional knowledge transfer across geographic contexts

The gains observed in locations with sparse POI observations raise the question of how functional alignment remains effective when local functional supervision is limited. One possibility is that physical–functional relationships learned from POIs elsewhere can transfer to locations with fewer local observations. We therefore examined whether functional knowledge learned through POI alignment remained informative as supervision was progressively shifted away from the target location (Fig. 4a). We first varied POI availability across three geographic scopes: target city (City), country including the target city (Country), and Country Leave-One-City-Out (Country LOCO), which excluded all target-city POIs from the country pool. We then conducted a fully crossed experiment in which representations trained on POIs from each of fifteen source countries were evaluated in all fifteen target cities.

Downstream performance generally improved as more POIs were retained across all three geographic scopes (Fig. 4b). With only 1% of each pool retained, Country and Country LOCO supervision already exceeded the AEF baseline of 0.777 (mean $R^2$ of 0.793 and 0.790, respectively), whereas City supervision required more POIs to match the baseline. Gains generally showed diminishing returns as additional POIs were introduced. At low retention percentages, Country and Country LOCO outperformed City, reflecting their larger absolute POI pools and greater exposure to transferable samples. In contrast, City achieved the highest performance with full POI availability, suggesting that local supervision may also contain location-specific functional signals that become diluted when training is expanded to broader geographic pools.

Country LOCO provided a rigorous evaluation of transfer in the complete absence of local functional supervision. Country-level alignment with all available POIs improved mean $R^2$ over AEF by 0.043. After every target-city POI was removed, the gain remained 0.038 (Fig. 4b), retaining 87.9% of the improvement obtained under Country training. Much of the benefit of functional alignment therefore did not require direct exposure to POIs from the target city, indicating that physical–functional relationships learned elsewhere remained informative for cities contributing no local POI supervision. Task-specific retention curves and city-level full-pool results are reported in Supplementary Fig. 2 and Supplementary Table 5.

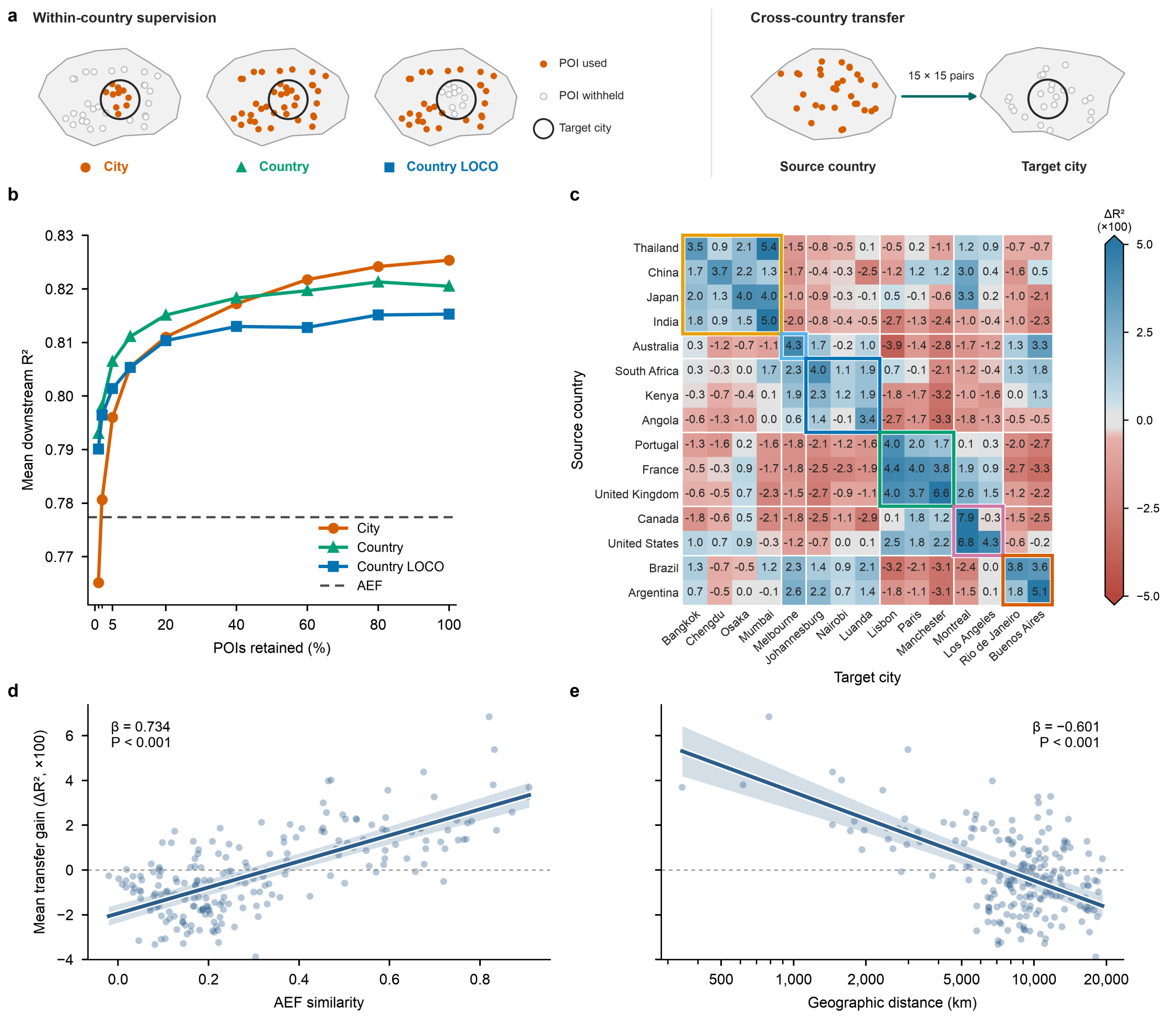

**Fig. 4. Transferability of urban functional knowledge across geographic contexts. a,** Experimental design. Left, within-country supervision: PURE is trained on POIs from the target city only (City), from the whole country including the target city (Country), or from the rest of the country with all target-city POIs withheld (Country LOCO). Right, cross-country transfer: PURE is trained on all POIs of one source country and applied to a target city in another country. Orange dots, POIs used for training; open circles, POIs withheld. **b,** Five-task mean downstream $R^2$ under City, Country and Country LOCO supervision across POI retention levels; the dashed line indicates the AEF baseline. **c,** Mean transfer gain in $R^2$ relative to AEF ($\Delta R^2 \times 100$) for the fully crossed design of 15 source countries and 15 target cities. Diagonal cells are matched source-country–target-city combinations; off-diagonal cells represent cross-country transfer. Coloured blocks group source countries and target cities on the same continent; the colour scale is truncated at ±5. d–**e,** Cross-country transfer gain against AEF representation similarity (**d**) and geographic distance (**e**) for the 210 off-diagonal pairs. Lines show marginal fits from separate linear mixed-effects models with 95% confidence intervals; $\beta$ is the standardised coefficient with its two-sided Wald $P$ value. Geographic distance is shown on a logarithmic axis and was log-transformed for modelling. Gains are averaged across five tasks and five random seeds.

The geographic scope and constraints of functional knowledge transfer were further examined across countries (Fig. 4c). All fifteen matched source-country–target-city combinations outperformed AEF, with a mean $R^2$ gain of 0.043. Among the 210 off-diagonal cross-country transfers, 88 (42%) outperformed AEF, with a mean gain of −0.001 and a median gain of −0.003. Positive gains were not randomly distributed but clustered along the matrix diagonal and among source–target pairs within the same continent. Of the 28 cross-country pairs within a continent, 26 improved on AEF, compared with 62 of the 182 pairs across continents (Figs. 1b and 4c). The representation trained in the United States, for example, improved mean $R^2$ in Montreal by 0.068, while transfers from Thailand to Mumbai and from France to Lisbon yielded gains of 0.054 and 0.044, respectively. The near-zero cross-country mean therefore concealed substantial variation among source–target combinations, showing that transfer was spatially structured rather than uniformly beneficial across geographic contexts.

Transfer gain increased with AEF representation similarity ($\beta = 0.734$, $P < 0.001$; Fig. 4d) and decreased with geographic distance ($\beta = -0.601$, $P < 0.001$; Fig. 4e). Separate mixed-effects models included crossed random intercepts for source country and target city. These patterns indicate that functional knowledge learned from POI observations can transfer beyond the locations in which it was learned, providing a mechanism through which functional alignment can benefit targets with limited local POI supervision. Transfer was nevertheless geographically structured, with stronger gains between physically similar and geographically closer source–target contexts. Removing geographic positional encoding left the source–target transfer matrix nearly unchanged (Pearson $r = 0.970$), indicating that this structure was not explained simply by direct encoding of absolute location (Supplementary Fig. 3 and Supplementary Table 6).

# Discussion

Globally consistent Earth representations capture substantial variation for urban analysis, but physical characteristics alone provide incomplete information about how places are organised and used. By aligning POI-derived functional semantics with AEF, PURE improved prediction across all five global tasks, indicating that functional observations add information beyond physical form. Similar gains were observed across more fine-grained downstream tasks in the supplementary evaluations (Supplementary Information, Section 3). The largest absolute gains, for NTL and GDP, suggest that functional semantics add information about economic activity and urban use that is only partly expressed in physical form. For POP, $PM_{2.5}$ and LST, functional alignment provided additional explanatory information relevant to human activity against relatively strong AEF baselines, and for $PM_{2.5}$ it removed the largest share of unexplained variance.

Functional knowledge learned from well-observed locations can remain informative where local observations are sparse. Whereas POIs typically describe the locations where they are recorded, PURE uses them as supervision for learning physical–functional relationships that can be applied elsewhere. When all target-city POIs were excluded, 87.9% of the gain obtained from country-level alignment was retained, showing that much of the learned functional knowledge could remain useful without direct local supervision. Cross-country experiments extended this finding beyond the national context. Knowledge learned from POIs in one country could remain useful for cities in previously unseen countries. However, such transfer was not universal. Gains were generally stronger within continents and were associated with greater physical similarity and shorter geographic distance between source and target contexts. Similar physical environments may preserve relationships between urban form and function, while geographic proximity may capture shared socioeconomic, cultural and development contexts that shape how those environments are used. As these contexts diverge, the same physical–functional relationships may become less transferable.

The transferability of urban functional knowledge offers a way to mitigate the consequences of persistent inequalities in urban data availability. Although detailed functional data remain concentrated in better-observed cities and regions, knowledge derived from these observations can have value beyond their geography of collection. This creates a pathway for comparative urban analysis to make broader use of existing data without requiring equally dense observations everywhere. The growing availability of globally consistent representations, together with methods for transferring knowledge through them, may provide one route towards this goal. Such approaches cannot substitute for local evidence, but they can reduce the extent to which analytical capacity is determined by where data happen to be abundant.

Several limitations define the scope of these conclusions. POI data are shaped by platform coverage, mapping practices, category systems, language and the uneven visibility of activities. PURE also inherits uncertainties from AEF and downstream reference datasets, including differences in target resolution, spatial smoothness and measurement uncertainty. The transfer experiments demonstrate representation-level transfer rather than label-free prediction, and associations with physical similarity and geographic distance should not be interpreted as causal. Finally, the fifteen source countries and target cities represent only a subset of global urban diversity. Future work may extend transfer tests across broader urban contexts, examine source selection and local calibration, and incorporate culturally specific functional semantics.

# Methods

## 1. Datasets

### 1.1 Data used to learn urban representations

PURE was developed by integrating a globally consistent physical representation of urban space with geographically discrete functional observations. The physical representation was obtained from AEF[20], which provides annual 64-dimensional embeddings of the global land surface at a native spatial resolution of 10 m. We used the 2024 embedding epoch and sampled the AEF embeddings onto the 100 m grid used throughout this study.

For each grid cell $i$ associated with POIs, the representation learning record was

$$\mathcal{D}_i = \left(\mathrm{a}_{i,0}, a_{i,1}, t_i, x_i, y_i\right), \tag{1}$$

where $\mathrm{a}_{i,0}$ is the AEF feature of the centre cell, $a_{i,1}$ is the mean AEF feature over valid pixels in a centred 3 × 3 local window, $t_i$ is the embedded description of the associated POI, and $(x_i, y_i)$ is the position of the cell. The two AEF views provided information about local physical consistency, whereas the embedded POI description represented POI-derived functional semantics and place identity.

POIs outside China were obtained from the Overture Maps Foundation Places dataset, release 2024-12-18.0. Overture Places records include point coordinates, names and category fields that were used to generate POI textual descriptions. POIs in China were obtained from AMAP. All POI descriptions followed the same English-language template:

*"A place of {category_leaf}, a type of {category_top}, named {name}."*

Descriptions were encoded using Qwen/Qwen3-Embedding-8B[42], and the first 512 dimensions of each 4,096-dimensional Matryoshka embedding were retained and L2-normalised as compact text features. The resulting training dataset comprised 84,618,817 spatially aligned records of AEF features, POI-derived functional semantic features and geographic positions.

### 1.2 Downstream datasets

We evaluated AEF and PURE on global versions of five CityRep benchmark tasks[43], representing demographic, economic, activity and human-environmental conditions: POP, GDP, NTL, $PM_{2.5}$ and LST. Dataset versions, years and spatial resolutions are summarised in Table 1.

Table 1. Global downstream benchmark tasks.

| Task | Prediction target | Source and year | Spatial support |
|---|---|---|---|
| POP | Population count | WorldPop constrained population counts, R2024A, 2024[44] | 3 arc-second raster, approximately 100 m at the equator |
| GDP | Total economic output (PPP) | Kummu et al. updated gridded GDP dataset, 2024[45] | 30 arc-second raster |
| NTL | Mean masked nighttime radiance | VIIRS Nighttime Lights Annual V2.2, 2024[46] | Approximately 15 arc-second / 500 m raster |
| $PM_{2.5}$ | Annual mean ground-level $PM_{2.5}$ concentration | SEDAC / ACAG SatPM V5.GL.04, 2022[47] | 0.01-degree raster |
| LST | Annual mean daytime LST | MODIS/Terra MOD11A2.061, 2024[48] | 1 km source raster |
| Urban boundary | Included settlement classes | GHS-SMOD R2023A V2.0, epoch 2025[49] | 1 km categorical grid |

The evaluation domain was restricted to settlement locations using GHS-SMOD R2023A V2.0, epoch 2025, classes 13, 21, 22, 23 and 30[49]. The 1 km categorical mask was reprojected by nearest neighbour and membership was evaluated at each target-cell centre, without connected-component filtering. For targets coarser than 100 m, AEF and PURE features were aggregated to the target support using overlap-area-weighted means and then L2-normalised. The same spatial support and aggregation procedure were used for both representations.

## 2. Learning physical–functional relationships through POI-guided alignment

### 2.1 Representation model

The model contained separate physical and functional pathways (the AEF and text encoders in Fig. 1a) during representation learning. At inference, the physical pathway transformed the 64-dimensional AEF feature into the 128-dimensional PURE representation $z_i$ at pixel-centre geographic location $l_i$:

$$z_i = f_\theta(a_{i,0}, l_i), \qquad \|z_i\|_2 = 1. \tag{2}$$

The physical pathway first combined a nonlinear projection of the AEF feature with a gated residual shortcut in a 128-dimensional space. The result was projected into a hidden space of width 2,048, where geographic position was incorporated as described below, before being mapped to the final 128-dimensional representation. The functional pathway used a linear projection to map the 512-dimensional POI-derived functional semantic feature into the same 128-dimensional space for contrastive learning. Outputs from both pathways were L2-normalised before similarity was calculated.

During training, each grid-cell representation was aligned with the representation of its associated POI description and separated from unmatched descriptions in the batch. After training, the functional pathway was removed, and PURE was generated using only the physical pathway, with AEF features and geographic position as inputs.

## 2.2 Geographic position conditioning

Because similar physical forms may have different meanings in different geographic settings, the physical pathway incorporated broad positional context. The stored grid indices were converted to pixel-centre latitude $\phi_i$ and longitude $\lambda_i$, which were encoded using real spherical harmonics of degree 8:

$$p_i = \mathrm{PE}_{\mathrm{SH}}(\phi_i, \lambda_i), \qquad p_i \in \mathbb{R}^{81}, \tag{3}$$

where SH denotes the spherical harmonics function. The physical and positional features were projected separately and combined through scaled addition:

$$\begin{gathered} \tilde{h}_i = \mathrm{LayerNorm}\big(h_i^{\mathrm{AEF}}\big) + \rho\,\mathrm{LayerNorm}\big(h_i^{\mathrm{pos}}\big), \\ z_i = \mathrm{Norm}(W\tilde{h}_i), \end{gathered} \tag{4}$$

where $\mathrm{LayerNorm}(\cdot)$ denotes layer normalisation, $W$ is the final projection and $\rho$ is a learnable scalar controlling the contribution of geographic position. The value of $\rho$ was initialised to 0.001. The initial model was therefore close to a position-independent transformation and incorporated geographic information only when it contributed to the representation learning objective.

## 2.3 Training objectives and optimisation

The training of PURE combined functional alignment, local physical consistency and preservation of the original AEF relational structure. The two alignment terms used a symmetric contrastive objective based on InfoNCE and modern contrastive representation learning methods[50,51]. Let $Z_0$ denote the centre-cell representations derived from the AEF, $Z_{\mathrm{POI}}$ the projected functional-semantic representations, and $y$ the identity target over rows in a training batch. The symmetric AEF-POI loss was

$$\mathcal{L}_{\mathrm{AEF-POI}} = \frac{1}{2}\mathrm{CE}\left(\frac{Z_0 {Z_{\mathrm{POI}}}^{\top}}{\tau_{\mathrm{POI}}}, y\right) + \frac{1}{2}\mathrm{CE}\left(\frac{Z_{\mathrm{POI}} {Z_0}^{\top}}{\tau_{\mathrm{POI}}}, y\right), \tag{5}$$

where $\mathrm{CE}$ denotes the cross-entropy loss, $\tau$ is the temperature parameter, and $y$ contains the indices of the matched AEF–POI pairs within each batch, such that the diagonal entries of the cross-modal similarity matrix are treated as positive pairs.

A second symmetric loss aligned the centre-cell representation $Z_0$ with the local-window AEF view $Z_1$:

$$\mathcal{L}_{\mathrm{AEF-AEF}} = \frac{1}{2}\mathrm{CE}\left(\frac{Z_1 {Z_0}^{\top}}{\tau_{\mathrm{AEF}}}, y\right) + \frac{1}{2}\mathrm{CE}\left(\frac{Z_0 {Z_1}^{\top}}{\tau_{\mathrm{AEF}}}, y\right). \tag{6}$$

To preserve the relational structure of the original AEF embedding space, the 64-dimensional AEF features were additionally used as a frozen relational teacher, following the broader idea of knowledge distillation[52] and its relational form[53]. For a subset $\mathcal{I}$ of up to 512 rows sampled from each batch, pairwise cosine similarities of the original AEF and current centre-cell encoder outputs were converted into row-wise neighbourhood distributions:

$$P_{ij}^{T} = \frac{\exp\left(\frac{\bar{a}_{i,0}^{\mathsf{T}}\bar{a}_{j,0}}{\tau_{\mathrm{rel}}}\right)}{\sum_{k\in\mathcal{I},k\neq i}\exp\left(\frac{\bar{a}_{i,0}^{\mathsf{T}}\bar{a}_{k,0}}{\tau_{\mathrm{rel}}}\right)}, i\neq j,$$
$$P_{ij}^{S} = \frac{\exp\left(\frac{\bar{z}_{i,0}^{\mathsf{T}}\bar{z}_{j,0}}{\tau_{\mathrm{rel}}}\right)}{\sum_{k\in\mathcal{I},k\neq i}\exp\left(\frac{\bar{z}_{i,0}^{\mathsf{T}}\bar{z}_{k,0}}{\tau_{\mathrm{rel}}}\right)}, i\neq j, \tag{7}$$

where the bars denote L2-normalised vectors and self-similarities were excluded. For each reference sample $i$, $P_{ij}$ is the relative probability assigned to sample $j$ as its neighbour within the selected subset. The original AEF features defined the teacher distribution, whereas the current centre-cell encoder outputs defined the student distribution. The relational-distillation loss was

$$L_{\mathrm{rel}} = \frac{1}{|\mathcal{I}|}\sum_{i\in\mathcal{I}} D_{\mathrm{KL}}(P_i^{T} \parallel P_i^{S}). \tag{8}$$

This objective encouraged the student encoder to preserve the within-batch neighbourhood geometry of the original AEF representation. The complete training objective was

$$L = \alpha L_{\mathrm{AEF-AEF}} + (1-\alpha)L_{\mathrm{AEF-POI}} + \beta L_{\mathrm{rel}}. \tag{9}$$

We set $\alpha = 0.2$, $\beta = 0.5$, $\tau_{\mathrm{POI}} = \tau_{\mathrm{AEF}} = 0.07$, and $\tau_{\mathrm{rel}} = 0.1$. The model was optimised with AdamW using a learning rate of $10^{-4}$, weight decay of $10^{-4}$, a batch size of 2,048, and seed 45. Training lasted 10 epochs with a fixed learning rate, and the checkpoint at epoch 10 was used in the main analyses.

After training, the functional pathway and relational-distillation objective were discarded. All global inference, downstream prediction, and released PURE features were generated using only AEF features and geographic position.

### 2.4 Spatially structured sampling

Training batches were constructed to represent both local urban context and broader geographic diversity. Around each sampled anchor cell, additional observations were drawn from square spatial rings corresponding approximately to distances of 10 km, 100 km and 1,000 km, together with observations from other global locations.

These groups contributed 0.50, 0.20, 0.20 and 0.10 of each batch, respectively. Nearby observations provided difficult contrasts among locations that could share similar physical or functional characteristics. More distant observations increased the diversity of urban functions and geographic conditions represented within each batch.

## 3. Evaluation design

### 3.1 Global downstream prediction

The global analysis tested whether the AEF representation and the functionally enriched PURE representation supported consistent measurement of urban and human-environmental conditions across geographic contexts.

The primary comparison was controlled at the level of downstream data and modelling. AEF and PURE used the same target observations, spatial aggregation, training-validation-test partitions, prediction architecture, optimisation settings and performance metrics. POIs were not supplied to the downstream models. Differences in performance therefore reflected the information available in the two representations under an otherwise identical prediction protocol.

Within each macroregion-task unit, a separate MLP was trained for every represented country or territory. Each prediction head contained one hidden layer of width 1,024. Country-specific heads prevented countries with large numbers of grid cells from dominating a shared regional model and accommodated differences in national outcome distributions.

Downstream heads were trained for a maximum of 100 epochs using a learning rate of 0.001 and early stopping patience of 10 epochs. Evaluations were repeated with five random seeds (42, 24, 7, 0 and 100). Batch size was 8,192 for POP and 128 for GDP, NTL, $PM_{2.5}$ and LST.

Targets were $z$-standardised using parameters estimated from the training subset. Observations were assigned to training, validation and test subsets using 1 km spatial blocks rather than random pixels, reducing direct information leakage between nearby observations arising from spatial autocorrelation.

Performance was assessed using the coefficient of determination ($R^2$) after inverse-transforming predictions to the original target units and mean absolute error (MAE) on targets standardised with training-subset statistics (Supplementary Table 4). $R^2$ was the primary reported metric. Country-level metrics were averaged across the five seeds. Reported uncertainty values are standard deviations across seed-specific country means. To express gains relative to the remaining error of the physical representation, we additionally calculated the proportional reduction in unexplained variance:

$$\mathrm{UER} = \frac{(1 - R^2_{\mathrm{AEF}}) - (1 - R^2_{\mathrm{PURE}})}{1 - R^2_{\mathrm{AEF}}} \times 100. \tag{10}$$

### 3.2 Dependence on POI availability

We next examined whether PURE gains were concentrated in countries with high POI availability. For each country and task, performance gain was defined as

$$\Delta R^2_{c,t} = R^2_{\mathrm{PURE},c,t} - R^2_{\mathrm{AEF},c,t}, \tag{11}$$

where $c$ denotes country or territory and $t$ denotes task.

The geographic distribution of $\Delta R^2$ was first summarised across continents. Country-level analyses then evaluated whether gains varied systematically with POI availability. POI availability was measured as the number of POIs per 1,000 residents, using population totals aggregated from the 2024 WorldPop raster.

A fixed cross-task cohort was defined using source support. Countries and territories were retained when the POP task contained more than 10,000 valid settlement target cells and each of GDP, NTL, $PM_{2.5}$ and LST contained more than 1,000. The intersection contained 142 countries and territories and was used for all global performance comparisons (Figs. 1a and 2). For Fig. 3, country-level variability, continent gain summaries and density associations used all 142 cohort members with matched POI and population statistics.

For each task, country $R^2$ values were first averaged across the five seeds and $\Delta R^2$ was regressed on $\log_2$-transformed POIs per 1,000 residents using ordinary least squares, so that slopes express the change in $\Delta R^2$ per doubling of POI density. Cross-country performance variability was summarised with the median absolute deviation of country-level $R^2$ to reduce sensitivity to extreme values.

Continent summaries were calculated from the eligible country-level values, giving each country equal weight within a task. Reported slopes and two-sided P values were obtained from ordinary least squares fits to the full, unclipped values with no observations excluded as outliers.

### 3.3 Transferability analysis

We examined transferability by varying the amount and geographic scope of POI supervision and by testing whether functional knowledge learned in one geographic context remained useful in another. Fifteen metropolitan areas were selected for within-city evaluation: Bangkok, Johannesburg, Los Angeles, Melbourne, Montreal, Mumbai, Osaka, Paris, Rio de Janeiro, Chengdu, Manchester, Nairobi, Lisbon, Luanda and Buenos Aires, representing fifteen countries across six continents.

To examine the effect of POI availability on the representation, PURE was trained using 1%, 2%, 5%, 10%, 20%, 40%, 60%, 80% or 100% of POIs under three geographic scopes. City used POIs within the target-city support. Country used POIs from the entire country containing the target city. Country LOCO used the same country-level pool after excluding all target-city POIs, allowing non-local supervision to increase while local supervision remained absent. Subsets were nested using a fixed seed. Retention percentages were defined relative to each pool rather than matched by absolute POI count. Changes within each curve therefore characterise responses to increasing POI availability, whereas differences among scopes also reflect variation in absolute POI quantity, geographic coverage and training exposure. AEF was included as the zero-POI baseline.

Cross-country transfer was evaluated using a fully crossed 15 × 15 source–target design. A separate PURE representation was trained using all available POIs from each source country and evaluated in every target city. Diagonal entries represented matched source-country and target-city combinations, whereas off-diagonal entries represented cross-country transfer. For each source–target evaluation, the trained representation was frozen, and a new downstream predictor was fitted using target-city labels. The analysis therefore measured representation-level transfer rather than zero-shot transfer of the downstream predictor. Evaluation covered GDP, POP, NTL, $PM_{2.5}$ and LST, using identical target observations, spatial partitions, prediction architectures and downstream random seeds.

To characterise variation among the 210 off-diagonal transfers, AEF representation similarity, used as a measure of physical similarity, was calculated from the original 64-dimensional AEF representation. For each source country, AEF sampling covered all settlements within its national boundary, and for each target city it covered the full predefined settlement support. The settlement mask followed the main evaluation protocol. Locations were mapped to 5,000 unique valid native 100 m AEF cells per country or city, with cell centres required to remain inside the settlement mask and corresponding geographic boundary. Fixed random seeds were used, and sampling was independent of POI locations and downstream labels. The sampled 64-dimensional features were averaged and L2-normalised before cosine similarity was calculated. Transfer gain was defined as the difference in $R^2$ between source-trained PURE and AEF for the same target city and task and was averaged across five tasks and five random seeds. Geographic separation was measured as the great-circle distance from the source-country capital to the centroid of the target city's predefined settlement support. Distance was log-transformed before modelling. Associations between transfer gain and physical similarity or geographic distance were evaluated using separate linear mixed-effects models with crossed random intercepts for source country and target city. Both predictors and transfer gain were standardised, and two-sided Wald tests were used for the fixed-effect coefficients. The reported coefficients represent separate associations rather than mutually adjusted effects.

All PURE variants in the transferability analysis used the same architecture and training objectives as the main model and were trained from scratch for 10 epochs with a batch size of 512, fixed learning rate $10^{-4}$ with no decay and seed 45.

### 4. Global dataset generation and release

The global PURE dataset was generated using the 2024 annual AEF embeddings. PURE representations were produced for grid cells with valid AEF features at a nominal spatial resolution of 100 m. The resulting coverage spans global land areas, including small islands, while Antarctica is excluded from the public release.

Each grid cell was processed through the trained AEF encoder to produce a 128-dimensional PURE embedding. This procedure uses the physical–functional relationships learned during POI alignment to generate PURE representations at locations where AEF features are available. POIs are used during model training but are not required to generate or use the released embeddings.

The dataset release is designed as georeferenced GeoTIFF files organised by country and first-level administrative region. Embeddings are stored using 8-bit quantisation, with the corresponding decoding parameters provided to recover the original values.

## Data availability

The released PURE embeddings are available at https://spacetimeai.com/pure/ and https://archive.org/details/poi-aligned-urban-embedding-v1-index. The AlphaEarth Foundations Satellite Embedding dataset is produced by Google DeepMind and available through Google Earth Engine (https://developers.google.com/earth-engine/datasets/catalog/GOOGLE_SATELLITE_EMBEDDING_V1_ANNUAL). Overture Places release 2024-12-18.0 is available from the Overture Maps Foundation (https://overturemaps.org). WorldPop, GDP, VIIRS nighttime-lights, $PM_{2.5}$, MODIS LST and GHS settlement data are available from the repositories cited in Methods. The original AMAP POI records used for China are not redistributed.

## Code availability

All code used to produce the results reported in this study is available on GitHub at https://github.com/inwind0212/PURE.

## Acknowledgements

This work was supported by UCL CEGE Scholarship, UCL Dean's Prize, National Natural Science Foundation of China (Grant No. 42301485), China Scholarship Council (Grant No. 202106270039 and 202406270207), Engineering and Physical Sciences Research Council (EPSRC) Industrial CASE Studentship with Ordnance Survey (Grant No. EP/W522077/1) and UCL Research Excellence Scholarship.

# Supplementary Information for: Urban Functional Knowledge Transfer through Global Representations amid Uneven Data

**Junyuan Liu[1,*], Xinglei Wang[1,*], Guangsheng Dong[2,1,†], Hao Sha[2], Zichao Zeng[1], Zhicheng Deng[1], Jiazhuang Feng[1], Xiang Wang[2], and Tao Cheng[1,‡]**

[1]SpaceTimeLab, Department of Civil, Environmental and Geomatic Engineering, University College London, London, UK

[2]State Key Laboratory of Information Engineering in Surveying, Mapping and Remote Sensing, Wuhan University, Wuhan, China

*These authors contributed equally.
† Email: guangshengdong@whu.edu.cn
‡ Email: tao.cheng@ucl.ac.uk

# 1 Supplementary Overview

This Supplementary Information provides additional evidence and methodological details supporting the global evaluation of POI-aligned Urban Embedding (PURE). Section 2 extends the main global benchmark, which evaluates PURE across five globally harmonised gridded targets, population (POP), gross domestic product (GDP), nighttime lights (NTL), $PM_{2.5}$ and land surface temperature (LST), in 142 countries and territories. It documents the POI corpus and regression sample support, examines the geographic distribution of the performance gain of PURE under uneven POI availability, resolves knowledge-transfer performance at the city level, and reports additional robustness analyses.

Section 3 provides complementary downstream validation using data and prediction settings that differ from the global raster benchmark. While the five global tasks offer broad and geographically consistent coverage, they are all based on gridded products and share a common cell-level regression formulation. We further evaluate PURE in more focused tasks with prediction targets being (1) the official small-area socioeconomic indicators from the United States and England, and (2) Census tract-level commuting flow records from the United States. These tasks introduce different data sources, spatial supports and prediction structures, including both area-level attributes and directed interactions between places. These two downstream evaluations are not intended simply to demonstrate additional applications of PURE. Rather, they test whether the benefits of functional enrichment persist beyond the global raster benchmark, providing complementary evidence for the transferability and reusability of the learned representation.

# 2 Additional Global Evaluation and Knowledge Transfer Analyses

## 2.1 Global POI corpus

The representation-learning corpus contained 84,618,817 POI-aligned records. Each record linked a standardised POI description and location to its 64-dimensional AlphaEarth Foundations (AEF) representation[1]. POIs outside China came from Overture Maps Places release 2024-12-18.0. POIs in China came from AMAP and used the same English description template. Qwen3-Embedding-8B[2] encoded the descriptions as 512-dimensional vectors. POI records, counts and density variables were used only for representation learning, not by downstream prediction heads.

Of these records, 84,592,091 POIs are mapped to 236 country or territory units with a 2024 WorldPop population denominator. The remaining 26,726 records contributed to representation learning but not to per-capita estimates. Asia contained the most POIs, North America had the highest availability per 1,000 residents, and Africa had the lowest. These six reporting regions did not determine the model architecture or downstream splits (Supplementary Table 1).

## 2.2 Geographic distribution of PURE gains under uneven POI availability

Global average performance can obscure whether representation gains are broadly distributed across geographic contexts or concentrated in particular regions or data-rich settings. This distinction is especially relevant for PURE because POI availability varies substantially across countries. We therefore examined the distribution of PURE gains from two complementary perspectives: their geographic consistency across continents and countries, and their relationship with differences in POI density.

We first examined whether the global improvement of PURE was broadly distributed across geographic regions and individual countries. Continent-level estimates used the same 142 evaluation units, with countries weighted equally within each task and split. Standard deviations reflect variation across five random seeds, and Russia was assigned to Europe. PURE increased mean $R^2$ in all 30 continent–task combinations (Supplementary Table 3). Five-task gains ranged from +0.018 in Latin America and the Caribbean to +0.058 in Europe.

**Supplementary Table 1: Regional coverage and density of the 2024 POI corpus.** "All POIs" includes the complete training corpus. "POIs with denominator" excludes records without a harmonised country or territory population estimate. Only matched records contribute to POIs per 1,000 residents. Occupied area is the summed WGS84 ground area of unique 100-m index cells containing at least one training POI. The final column uses all records and describes concentration within the observed footprint, not across the entire region.

| Region | Units | All POIs | POIs with denominator | POIs per 1,000 residents | Occupied area ($km^2$) | POIs per occupied $km^2$ |
|---|---|---|---|---|---|---|
| Asia | 53 | 44,261,062 | 44,261,062 | 9.24 | 112,547.81 | 393.26 |
| Europe | 50 | 14,859,895 | 14,859,895 | 19.94 | 46,134.65 | 322.10 |
| North America | 5 | 14,415,215 | 14,415,215 | 37.53 | 52,216.96 | 276.06 |
| Latin America and the Caribbean | 47 | 8,103,474 | 8,089,882 | 12.25 | 28,629.56 | 283.05 |
| Oceania | 24 | 1,186,863 | 1,186,863 | 25.91 | 4,274.93 | 277.63 |
| Africa | 57 | 1,792,308 | 1,779,174 | 1.19 | 6,131.96 | 292.29 |
| All regions | 236 | 84,618,817 | 84,592,091 | 10.41 | 249,935.87 | 338.56 |

**Supplementary Table 2: Target products and effective support in the global regression benchmark.** All support statistics refer to the fixed 142 countries and territories. Per-country ranges are minimum–maximum counts of unique valid target cells before the five repeated spatial splits.

| Task | Product (year) | Quantity used | Total cells | Median per unit | Range per unit |
|---|---|---|---|---|---|
| GDP | Global gridded GDP (2024) | Total GDP, 30 arcsec (band 8) | 5,445,433 | 10,544 | 1,611–1,062,581 |
| POP | WorldPop (2024) | Population count | 535,463,912 | 1,029,867 | 148,454–104,702,003 |
| NTL | VIIRS annual composite (2024) | Annual-average nighttime radiance | 21,785,054 | 42,147 | 6,323–4,248,966 |
| $PM_{2.5}$ | ACAG V5.GL.04 (2022) | Annual mean surface $PM_{2.5}$ | 3,771,580 | 7,317 | 1,083–736,539 |
| LST | MOD11A2 (2024) | Annual mean daytime LST | 5,413,820 | 10,358 | 1,559–1,059,639 |

The improvement was also widespread at the country level. PURE exceeded AEF in 678 of 710 country–task comparisons. Positive comparisons numbered 138/142 for GDP, 129/142 for POP, 131/142 for NTL, 141/142 for $PM_{2.5}$ and 139/142 for LST. Country-level PURE performance nevertheless retained substantial geographic variation across all five tasks (Supplementary Fig. 1). Together with the country-level gains mapped in Fig. 2b–f, these maps show that functional alignment broadly increased predictive performance without eliminating task-specific geographic structure. Thus, the global mean improvement was not driven by a small subset of continents or countries.

We next examined whether the magnitude or consistency of PURE gains depended on the availability of POI observations. Countries were first divided into quartiles according to POIs per 1,000 residents. Across POI-density quartiles ordered from lowest to highest, mean gains in $R^2$ were +0.029, +0.023, +0.064 and +0.036. PURE outperformed AEF in 172/180, 164/175, 171/175 and 171/180 country–task comparisons in the four quartiles, respectively. Improvements were therefore observed throughout the POI-density distribution, including the countries with the sparsest POI coverage.

## 2.3 Complementary evaluation using standardised MAE

The primary evaluation used $R^2$ to quantify how much variation in each downstream target was captured by the learned representations. As a complementary check, we also evaluated mean absolute error (MAE), which measures prediction error directly rather than relative explained variance. Because the five targets differ substantially in their units and numerical scales, targets were standardised within each country and split using statistics estimated from the training subset, allowing MAE to be compared across tasks. This complementary metric tests whether the improvements observed in $R^2$ are also reflected in smaller absolute prediction errors rather than being specific to the variance-based evaluation. PURE reduced standardised MAE for all five tasks (Supplementary Table 4). The five-task mean fell from $0.224 \pm 0.001$ to $0.201 \pm 0.001$, a 10.1% reduction calculated from unrounded values. Task-specific reductions ranged from 4.1% for POP to 21.5% for $PM_{2.5}$.

**Supplementary Table 3: Continent-level performance.** Values are equal-country mean $R^2 \pm$ s.d. across five random seeds. Russia is assigned to Europe. $\Delta R^2$ is PURE minus AEF, calculated before rounding.

| Continent | Task | AEF $R^2$ | PURE $R^2$ | $\Delta R^2$ |
|---|---|---|---|---|
| Africa | GDP | $0.772 \pm 0.011$ | $0.813 \pm 0.009$ | $+0.041$ |
| | POP | $0.833 \pm 0.005$ | $0.853 \pm 0.003$ | $+0.019$ |
| | NTL | $0.738 \pm 0.010$ | $0.765 \pm 0.008$ | $+0.027$ |
| | $PM_{2.5}$ | $0.915 \pm 0.002$ | $0.941 \pm 0.001$ | $+0.026$ |
| | LST | $0.908 \pm 0.002$ | $0.917 \pm 0.002$ | $+0.009$ |
| | Five-task mean | $0.833 \pm 0.004$ | $0.858 \pm 0.002$ | $+0.024$ |
| Asia | GDP | $0.754 \pm 0.008$ | $0.809 \pm 0.009$ | $+0.055$ |
| | POP | $0.794 \pm 0.009$ | $0.812 \pm 0.007$ | $+0.018$ |
| | NTL | $0.516 \pm 0.231$ | $0.644 \pm 0.037$ | $+0.128$ |
| | $PM_{2.5}$ | $0.919 \pm 0.002$ | $0.948 \pm 0.001$ | $+0.029$ |
| | LST | $0.917 \pm 0.001$ | $0.929 \pm 0.001$ | $+0.011$ |
| | Five-task mean | $0.780 \pm 0.046$ | $0.828 \pm 0.007$ | $+0.048$ |
| Europe | GDP | $0.796 \pm 0.009$ | $0.852 \pm 0.007$ | $+0.056$ |
| | POP | $0.756 \pm 0.002$ | $0.779 \pm 0.002$ | $+0.023$ |
| | NTL | $0.546 \pm 0.030$ | $0.606 \pm 0.020$ | $+0.060$ |
| | $PM_{2.5}$ | $0.788 \pm 0.001$ | $0.887 \pm 0.002$ | $+0.099$ |
| | LST | $0.765 \pm 0.005$ | $0.815 \pm 0.003$ | $+0.050$ |
| | Five-task mean | $0.730 \pm 0.007$ | $0.788 \pm 0.005$ | $+0.058$ |
| North America | GDP | $0.817 \pm 0.005$ | $0.862 \pm 0.005$ | $+0.045$ |
| | POP | $0.792 \pm 0.007$ | $0.804 \pm 0.001$ | $+0.012$ |
| | NTL | $0.699 \pm 0.036$ | $0.716 \pm 0.041$ | $+0.016$ |
| | $PM_{2.5}$ | $0.901 \pm 0.002$ | $0.929 \pm 0.001$ | $+0.029$ |
| | LST | $0.940 \pm 0.003$ | $0.946 \pm 0.002$ | $+0.006$ |
| | Five-task mean | $0.830 \pm 0.007$ | $0.851 \pm 0.008$ | $+0.022$ |
| Latin America and the Caribbean | GDP | $0.829 \pm 0.010$ | $0.862 \pm 0.011$ | $+0.033$ |
| | POP | $0.837 \pm 0.004$ | $0.845 \pm 0.002$ | $+0.008$ |
| | NTL | $0.691 \pm 0.051$ | $0.700 \pm 0.057$ | $+0.009$ |
| | $PM_{2.5}$ | $0.883 \pm 0.002$ | $0.913 \pm 0.003$ | $+0.029$ |
| | LST | $0.916 \pm 0.002$ | $0.925 \pm 0.002$ | $+0.009$ |
| | Five-task mean | $0.831 \pm 0.012$ | $0.849 \pm 0.013$ | $+0.018$ |
| Oceania | GDP | $0.810 \pm 0.030$ | $0.852 \pm 0.024$ | $+0.042$ |
| | POP | $0.625 \pm 0.021$ | $0.635 \pm 0.019$ | $+0.010$ |
| | NTL | $0.623 \pm 0.057$ | $0.658 \pm 0.051$ | $+0.035$ |
| | $PM_{2.5}$ | $0.912 \pm 0.005$ | $0.933 \pm 0.006$ | $+0.021$ |
| | LST | $0.951 \pm 0.002$ | $0.960 \pm 0.002$ | $+0.008$ |
| | Five-task mean | $0.784 \pm 0.020$ | $0.807 \pm 0.016$ | $+0.023$ |

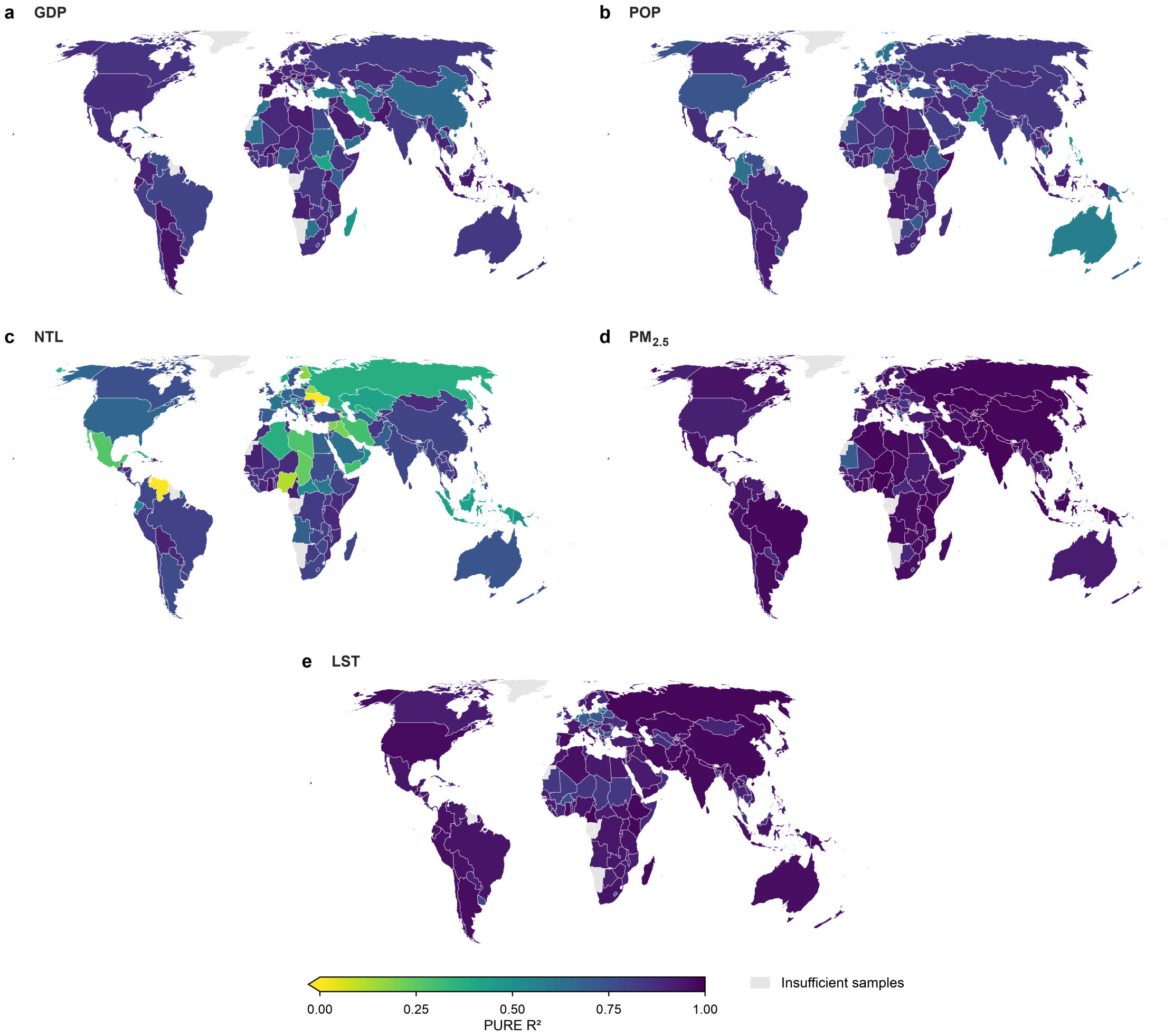

**Supplementary Figure 1: Country-level performance of PURE across the five global prediction tasks.** Mean PURE $R^2$ across five random seeds for **a,** GDP, **b,** POP, **c,** NTL, **d,** $PM_{2.5}$ and **e,** LST in the fixed cohort of 142 countries and territories. Values are shown on a common 0–1 colour scale; the corresponding gains over AEF are mapped in Fig. 2b–f. Countries and territories not meeting the task-specific sample threshold are shown in grey. Base map: Natural Earth.

**Supplementary Table 4: Standardised MAE for the primary global benchmark.** Values are equal-country means ± s.d. across five random seeds in the fixed set of 142 countries and territories. Targets were standardised separately for each country and seed using statistics estimated only from the training subset; AEF and PURE used identical statistics. The mean and percentage reductions were calculated from unrounded values. Lower values are better.

| Model | GDP | POP | NTL | $PM_{2.5}$ | LST | Mean |
|---|---|---|---|---|---|---|
| AEF | $0.189 \pm 0.001$ | $0.219 \pm 0.001$ | $0.225 \pm 0.003$ | $0.241 \pm 0.000$ | $0.244 \pm 0.000$ | $0.224 \pm 0.001$ |
| PURE | $\mathbf{0.168 \pm 0.001}$ | $\mathbf{0.210 \pm 0.001}$ | $\mathbf{0.212 \pm 0.002}$ | $\mathbf{0.189 \pm 0.000}$ | $\mathbf{0.226 \pm 0.000}$ | $\mathbf{0.201 \pm 0.001}$ |
| Reduction in MAE (%) | 11.3 | 4.1 | 5.7 | 21.5 | 7.4 | 10.1 |

## 2.4 Detailed POI-retention results

Supplementary Fig. 2 resolves the aggregate POI-retention analysis in Fig. 4b by downstream task. City used POIs within the target-city support, Country used all POIs in the corresponding country, and Country LOCO excluded POIs within the target-city support. Retention levels of 1, 2, 5, 10, 20, 40, 60, 80 and 100% were defined separately within each training pool; consequently, equal retention percentages correspond to different absolute numbers of POIs across scopes. AEF provides the no-POI reference. All conditions used the same 15 target-city supports and downstream samples. Representation models were trained for 10 epochs with batch size 512, learning rate $10^{-4}$ and seed 45.

Across tasks, increasing POI retention generally improved downstream performance, although the magnitude and shape of the response varied by outcome and supervision scope (Supplementary Fig. 2). Broader Country and Country LOCO pools were most effective at low retention, whereas City supervision generally benefited more from high retention. At full retention, the equal-city five-task mean $R^2$ was 0.825 for City, 0.821 for Country and 0.815 for Country LOCO, compared with 0.777 for AEF.

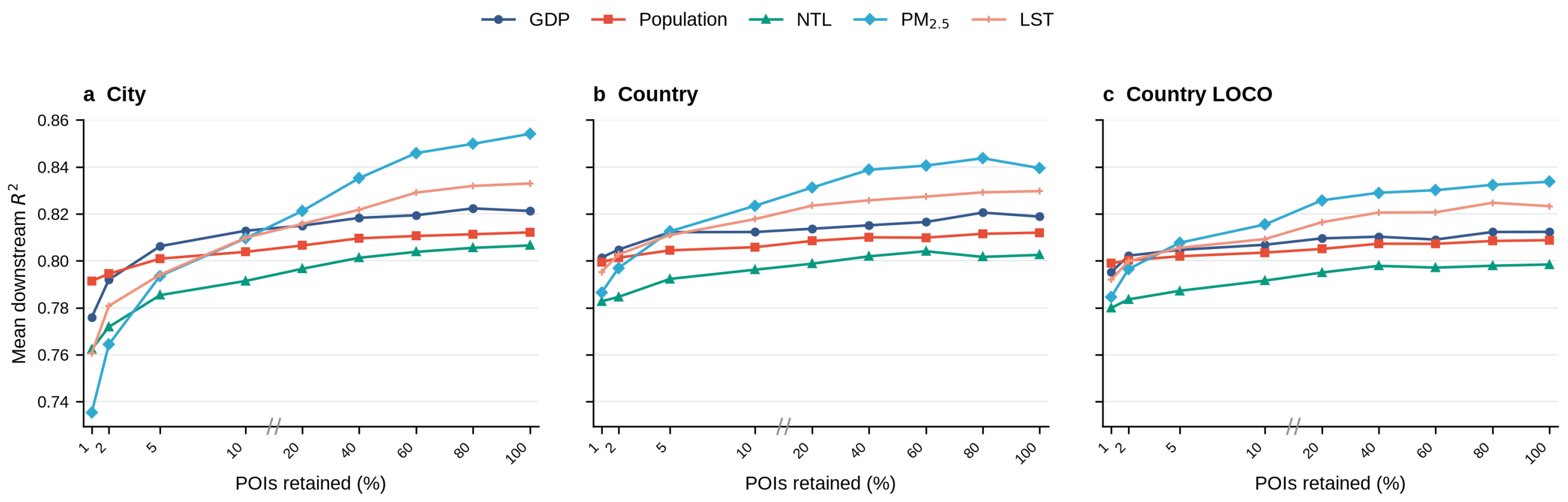

**Supplementary Figure 2: Task-specific responses to POI retention in 15 target cities.** Mean downstream $R^2$ for GDP, POP, NTL, $PM_{2.5}$ and LST under (a) City, (b) Country and (c) Country LOCO supervision. Values are averaged across 15 cities after averaging five random seeds. Horizontal positions are proportional to retention percentage within 1–10% and 20–100%, with different scales for the two intervals. The axis break between 10% and 20% indicates this change of scale; the connecting line across the break is a visual guide and its slope should not be interpreted quantitatively.

City-specific full-retention results are reported in Supplementary Table 5. The city-level results showed a broadly consistent pattern across geographic contexts. All three PURE supervision scopes outperformed AEF in every target city, indicating that the aggregate improvement was not driven by a small subset of locations. City achieved the highest mean $R^2$ in 13 of the 15 cities, while Country achieved the highest performance in Melbourne and Montreal. Among these 13 cities, Country ranked second in 12, with Lisbon as the only exception. Overall, these results show that full target-city supervision generally provided the strongest downstream performance, while country-level supervision typically ranked second and retained most of the improvement over AEF.

## 2.5 Additional evaluation and robustness analyses

### 2.5.1 Transfer without geographic positional encoding

Geographic positional encoding directly exposes absolute location to the model. It could therefore make nearby places more similar in the learned representation and create an apparent geographic structure in transfer without requiring transferable physical–functional relationships. To test this possibility, we repeated the complete 15-source by 15-target experiment after removing geographic positional encoding from PURE. The ablation was trained from scratch using the same 2024 POI feature store, architecture and objective weights, with the same 10-epoch schedule, batch size of 512, fixed learning rate of $10^{-4}$ and seed 45. Downstream target samples, spatial splits and five evaluation seeds were held fixed.

**Supplementary Table 5: City-specific performance with full POI availability.** Five-task mean $R^2$ after averaging five random seeds. Each PURE scope uses 100% of its respective POI pool. The final row reports the equal-city mean. Rankings and gains were determined from unrounded values.

| City | AEF | City | Country | Country LOCO |
|---|---|---|---|---|
| Bangkok | 0.862 | 0.906 | 0.898 | 0.892 |
| Chengdu | 0.772 | 0.825 | 0.809 | 0.806 |
| Osaka | 0.864 | 0.911 | 0.904 | 0.899 |
| Mumbai | 0.758 | 0.815 | 0.808 | 0.805 |
| Melbourne | 0.740 | 0.776 | 0.782 | 0.780 |
| Johannesburg | 0.765 | 0.811 | 0.805 | 0.795 |
| Nairobi | 0.837 | 0.849 | 0.849 | 0.845 |
| Luanda | 0.822 | 0.867 | 0.856 | 0.842 |
| Lisbon | 0.681 | 0.734 | 0.721 | 0.723 |
| Paris | 0.793 | 0.835 | 0.832 | 0.829 |
| Manchester | 0.728 | 0.794 | 0.794 | 0.792 |
| Montreal | 0.684 | 0.752 | 0.763 | 0.755 |
| Los Angeles | 0.769 | 0.823 | 0.812 | 0.807 |
| Rio de Janeiro | 0.804 | 0.844 | 0.842 | 0.839 |
| Buenos Aires | 0.782 | 0.839 | 0.833 | 0.819 |
| Mean | 0.777 | 0.825 | 0.821 | 0.815 |

Removing positional encoding preserved the source–target structure of the transfer matrix (Supplementary Fig. 3 and Supplementary Table 6). Across all 225 combinations, transfer gains from the two models were strongly correlated (Pearson $r = 0.970$). All 15 matched pairs remained positive, although their mean gain decreased modestly from 0.0431 to 0.0408. The 210 off-diagonal gains were similarly stable ($r = 0.961$), with an effectively unchanged mean of $-0.0013$ with positional encoding and $-0.0011$ without it.

Crucially, the geographic organisation of transfer remained intact. Within-continent cross-country gains remained strongly positive, changing only from 0.0232 to 0.0225, with 26 and 27 of 28 pairs positive, respectively. Between-continent gains remained weaker, changing from $-0.0051$ to $-0.0047$. Explicit positional information therefore made a small contribution in matched and within-continent contexts but was not responsible for the broader transfer structure. The stronger transfer among compatible geographic contexts observed in the main experiment cannot be explained simply by direct encoding of absolute location.

### 2.5.2 Sensitivity to the downstream prediction head

To assess whether the advantage of functional enrichment depends on a nonlinear downstream predictor, we replaced the default MLP with a country-specific linear head while keeping the 2024 AEF and PURE representations, target cells, 142 countries and territories, spatial splits and five downstream seeds fixed. Each linear head used training-only target standardisation and the same country-specific evaluation protocol as the main benchmark. The AEF linear results were reused after verifying identical AEF features, target and label files, splits and fitting parameters for all region–task jobs.

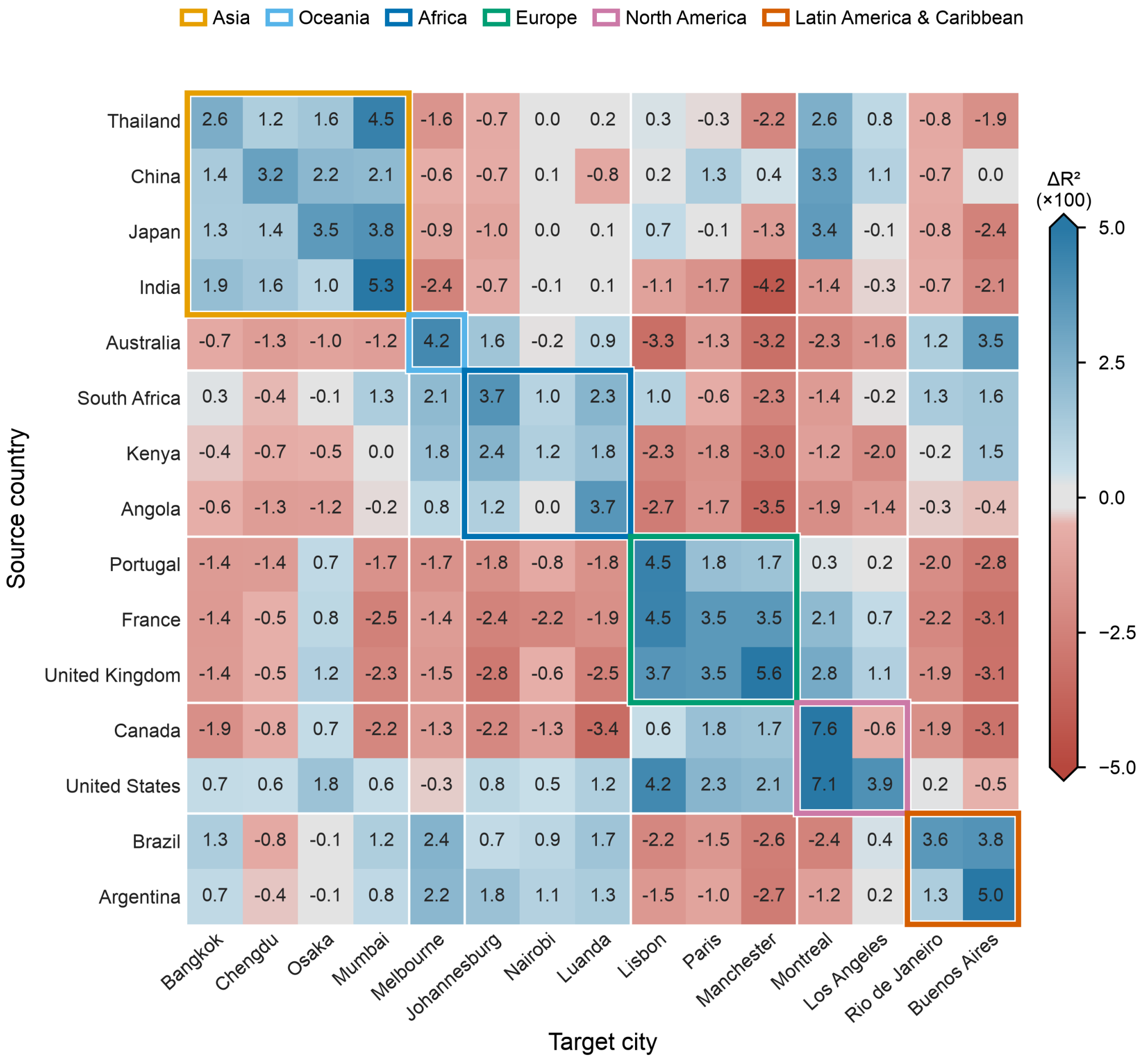

**Supplementary Figure 3: Cross-country transfer without geographic positional encoding.** Each cell shows mean $\Delta R^2$ $(\times 100)$ relative to the target-specific AEF baseline for a representation trained with all POIs from the source country and evaluated in the target city. Values are averaged across five tasks and five downstream seeds. Diagonal cells are matched source–target contexts; off-diagonal cells are cross-country transfers. Coloured blocks group source countries and target cities on the same continent; the colour scale is truncated at $\pm 5$. The experiment matches the main transfer analysis in Fig. 4c except that geographic positional encoding was removed during representation training.

**Supplementary Table 6: Cross-country transfer with and without geographic positional encoding.** Mean $\Delta R^2$ is relative to the target-specific AEF baseline after averaging five tasks and five downstream seeds. Difference is the no-position mean minus the main-model mean. Positive pairs report the main model followed by the no-position model. Pearson correlations are calculated across paired source–target combinations within each group. Mean gains and differences are displayed to four decimal places.

| Pair group | Pairs | With position | Without position | Difference | Positive pairs | Pearson $r$ |
|---|---|---|---|---|---|---|
| Matched | 15 | +0.0431 | +0.0408 | −0.0024 | 15 / 15 | 0.960 |
| All cross-country | 210 | −0.0013 | −0.0011 | +0.0002 | 88 / 93 | 0.961 |
| Within-continent cross-country | 28 | +0.0232 | +0.0225 | −0.0007 | 26 / 27 | 0.972 |
| Between-continent cross-country | 182 | −0.0051 | −0.0047 | +0.0004 | 62 / 66 | 0.943 |

PURE consistently outperformed AEF under both downstream prediction heads (Supplementary Table 7). With linear heads, PURE improved $R^2$ over AEF for all five tasks, with gains of $+0.099$ for GDP, $+0.060$ for POP, $+0.056$ for NTL, $+0.121$ for $PM_{2.5}$ and $+0.032$ for LST. The five-task mean increased from 0.602 for AEF to 0.675 for PURE, corresponding to $\Delta R^2 = +0.073$. The same pattern was observed with the default MLP, for which PURE exceeded AEF on every task and increased the five-task mean from 0.792 to 0.830 ($\Delta R^2 = +0.038$).

The persistence of PURE's advantage under a linear head suggests that functional enrichment produces a representation that is more directly aligned with the downstream prediction tasks, such that additional predictive information can be accessed through a simple linear readout. Notably, the mean improvement over AEF was larger with the linear head ($\Delta R^2 = +0.073$) than with the nonlinear MLP ($+0.038$). The larger gain under the linear head is consistent with task-relevant information in PURE being more directly accessible through a simple linear readout. The MLP nevertheless achieved higher absolute $R^2$ for both representations, showing that nonlinear prediction provides additional modelling capacity. Together, these results suggest that PURE improves not only downstream predictive performance but also the accessibility of task-relevant information in the learned representation.

**Supplementary Table 7: Global performance with linear and nonlinear downstream heads.** Values are equal-country mean $R^2\pm$ s.d. across five random seeds in the 142 countries and territories. The same target cells and splits are used for both head types.

| Head | Representation | GDP | POP | NTL | $PM_{2.5}$ | LST | Mean |
|---|---|---|---|---|---|---|---|
| Linear | AEF | $0.535 \pm 0.003$ | $0.585 \pm 0.003$ | $0.506 \pm 0.005$ | $0.611 \pm 0.001$ | $0.772 \pm 0.001$ | $0.602 \pm 0.002$ |
| Linear | PURE | $0.634 \pm 0.003$ | $0.645 \pm 0.003$ | $0.562 \pm 0.005$ | $0.732 \pm 0.001$ | $0.804 \pm 0.001$ | $0.675 \pm 0.002$ |
| MLP | AEF | $0.783 \pm 0.004$ | $0.799 \pm 0.003$ | $0.619 \pm 0.059$ | $0.880 \pm 0.001$ | $0.878 \pm 0.001$ | $0.792 \pm 0.011$ |
| MLP | PURE | $0.831 \pm 0.006$ | $0.817 \pm 0.003$ | $0.679 \pm 0.009$ | $0.925 \pm 0.001$ | $0.898 \pm 0.000$ | $0.830 \pm 0.003$ |

### 2.5.3 Ablation analysis of representation pretraining objectives

We trained two objective ablations from scratch using the same 2024 POI feature store, 512-dimensional text vectors, architecture, spatial sampling, 10-epoch schedule, batch size of 2,048 and seed 45 as the full global PURE model. One set the AEF relational-structure weight to zero; the other set the local physical consistency weight to zero. All other objectives and weights were retained. We compared the epoch-10 representations using the same 142 countries and territories, target cells, country-specific downstream heads, 1-km spatial-block splits and five downstream seeds.

Full PURE achieved a five-task mean $R^2$ of 0.830, compared with 0.821 without AEF relational structure and 0.809 without local physical consistency (Supplementary Table 8). Both objectives therefore contributed to global downstream performance. The larger reduction after removing local physical consistency was particularly apparent for GDP and NTL; removing AEF relational structure also reduced performance across all five tasks.

**Supplementary Table 8: Global downstream performance of 2024 representation-objective variants.** Values are equal-country mean $R^2\pm$ s.d. across five downstream seeds for the 142 countries and territories. All variants use epoch-10 checkpoints, identical target cells, country-specific prediction heads and spatial splits. Bold values indicate the best variant within each column.

| Model | GDP | POP | NTL | $PM_{2.5}$ | LST | Mean |
|---|---|---|---|---|---|---|
| Full PURE | $\mathbf{0.831 \pm 0.006}$ | $\mathbf{0.817 \pm 0.003}$ | $\mathbf{0.679 \pm 0.009}$ | $\mathbf{0.925 \pm 0.001}$ | $\mathbf{0.898 \pm 0.000}$ | $\mathbf{0.830 \pm 0.003}$ |
| No AEF relational structure | $0.824 \pm 0.003$ | $0.810 \pm 0.003$ | $0.665 \pm 0.010$ | $0.915 \pm 0.001$ | $0.891 \pm 0.000$ | $0.821 \pm 0.002$ |
| No local physical consistency | $0.800 \pm 0.006$ | $0.805 \pm 0.003$ | $0.647 \pm 0.013$ | $0.910 \pm 0.001$ | $0.886 \pm 0.000$ | $0.809 \pm 0.004$ |

## 3 Additional Independent Downstream Validation

### 3.1 Rationale and common representation aggregation

The five global regression tasks described in the main text and Section 2 provide the principal evaluation of PURE under globally harmonised spatial coverage. We additionally evaluate PURE in two downstream settings using official socioeconomic statistics and commuting records. These experiments are not intended simply to expand the number of demonstrated applications, but to test whether the advantages of PURE persist under different data sources, spatial supports and prediction formulations.

Small-area socioeconomic prediction evaluates whether the learned representation can help recover attributes associated with official statistical areas, whereas commuting flow generation tests whether interactions between residential origins and workplace destinations can be estimated from the learned representations.

Following previous experiments, two pretrained representations were compared: the 64-dimensional AEF representation and the 128-dimensional PURE representation. Neither the socioeconomic indicators nor the commuting flow observations were used during representation pretraining, so both provide independently supervised tests of the information encoded by the representations. For each spatial unit, i.e., the statistical-area polygon used in the two evaluations, raster embeddings were aggregated in the following way: raster cells entirely contained within a polygon received unit weight, whereas boundary raster cells were weighted by the fraction of their footprint intersecting the statistical area. Identical spatial weights were applied to AEF and PURE to facilitate directly comparable area-level representations.

The task-specific datasets, prediction protocols, and evaluation procedures and results are described separately below.

### 3.2 Small-area socioeconomic prediction

#### 3.2.1 Motivation and prediction task

Prediction of neighbourhood socioeconomic conditions from remotely sensed and urban functional information has been used to assess the extent to which observable characteristics of places encode socially meaningful variation[3–5]. Related work in official statistics has similarly emphasised the estimation of socioeconomic conditions for geographically disaggregated small areas[6]. Inspired by these studies, we test whether the learned representations encode information relevant to neighbourhood socioeconomic conditions reported for official statistical areas. A separate regression model was fitted for each socioeconomic indicator and input feature set. The input was either the area-level AEF embedding, the area-level PURE embedding or a longitude-latitude coordinate baseline, and the output was the corresponding continuous socioeconomic indicator. All feature sets used the same observations, target definitions, model architecture and outer training and test partitions.

#### 3.2.2 Data and preprocessing

We constructed a small-area socioeconomic prediction benchmark covering metropolitan regions in the United States and England. The US study areas comprised Los Angeles, Chicago and New York City, with Census tracts used as the unit of analysis. The English study areas comprised Greater London, West Midlands/Birmingham and West Yorkshire/Leeds, with Middle Layer Super Output Areas (MSOAs) used as the corresponding neighbourhood-scale statistical unit. Census tracts are relatively stable small-area units maintained by the US Census Bureau, whereas MSOAs are designed for neighbourhood-level statistical reporting in England.

Before indicator-specific missing-value filtering, the benchmark contained 4,088 US Census tracts and 1,612 English MSOAs. The US sample comprised 1,099 tracts in Los Angeles, 791 in Chicago and 2,198 in New York City. The English sample comprised 963 MSOAs in Greater London, 352 in West Midlands/Birmingham and 297 in West Yorkshire/Leeds.

**Socioeconomic indicators.** US socioeconomic data were obtained from the 2020–2024 American Community Survey (ACS) five-year estimates[7] through Census Reporter[8]. Five tract-level socioeconomic indicators were included, covering household income, poverty, unemployment, educational attainment and housing tenure. Median household income was transformed using the natural logarithm, whereas the remaining indicators were expressed as percentages.

English socioeconomic data were obtained from the 2021 Census tables distributed through Nomis[9]. Five MSOA-level indicators were included, covering employment, economic inactivity, educational attainment, self-reported health and multidimensional household deprivation. The precise definitions, source variables or tables, transformations and available sample sizes for all indicators are reported in Supplementary Table 9.

**Supplementary Table 9: Official socioeconomic indicators used in the small-area prediction benchmark.** The US indicators were derived from the 2020–2024 ACS five-year estimates, and the English indicators were derived from the 2021 Census. The reported sample size, $n$, reflects indicator-specific data availability.

| Country | Indicator | Definition and reference period | $n$ |
|---|---|---|---|
| United States | Log median household income | Natural logarithm of ACS `B19013_001`; household income in 2024 inflation-adjusted US dollars | 4,032 |
| United States | Poverty rate | ACS `B17001_002` divided by `B17001_001`, multiplied by 100 | 4,083 |
| United States | Unemployment rate | ACS `B23025_005` divided by `B23025_003`, multiplied by 100 | 4,084 |
| United States | Bachelor's degree or higher | Sum of ACS `B15003_022`–`B15003_025`, divided by `B15003_001` and multiplied by 100 | 4,086 |
| United States | Renter-occupied households | ACS `B25003_003` divided by `B25003_001`, multiplied by 100 | 4,078 |
| England | Employment rate | Census 2021 Table TS066; employed residents aged 16 years and over divided by all residents aged 16 years and over | 1,612 |
| England | Economic inactivity rate | Census 2021 Table TS066; economically inactive residents divided by all residents aged 16 years and over | 1,612 |
| England | Level 4 qualification or higher | Census 2021 Table TS067; residents with Level 4 qualifications or higher divided by all residents included in the qualification classification | 1,612 |
| England | Bad or very bad health | Census 2021 Table TS037; usual residents reporting bad or very bad health divided by all usual residents | 1,612 |
| England | Households deprived in two or more dimensions | Census 2021 Table TS011; households deprived in two, three or four dimensions divided by all households | 1,612 |

**Spatial support.** Each observation was represented by its complete Census tract or MSOA polygon rather than by a centroid, point buffer or fixed-radius neighbourhood. 11 New York City tracts lacking valid pixels in at least one representation were excluded from all comparisons between feature sets. This ensured that the spatial support of the representation corresponded directly to that of the official socioeconomic indicator.

The median area of the statistical units was $0.31\,\mathrm{km}^2$ for US Census tracts and $1.53\,\mathrm{km}^2$ for English MSOAs. The median number of valid PURE pixels overlapping a polygon was 55 for US tracts and 313.5 for English MSOAs. We averaged all valid overlapping pixel embeddings, weighting each by the fraction of its pixel area inside the polygon. A pixel fully inside the polygon therefore received a weight of 1, whereas a pixel with half its area inside received a weight of 0.5. The sum of these weights gives the equivalent number of complete pixels contributing to the average. These sums had medians of 39.3 full-pixel equivalents per US tract and 248.5 per English MSOA. Region-specific polygon areas are reported in Supplementary Table 10, and the numbers of overlapping pixels and full-pixel equivalents are summarised in Supplementary Table 11. The weighting procedure and observed distributions are illustrated in Supplementary Fig. 4.

**Supplementary Table 10: Spatial support of the small-area socioeconomic prediction benchmark.** Area statistics refer to the complete Census tract or MSOA polygons used to associate socioeconomic indicators with the urban representations.

| Country | Study region | Spatial unit | $n$ | Median area ($km^2$) | 95th percentile area ($km^2$) |
|---|---|---|---|---|---|
| United States | Los Angeles | Census tract | 1,099 | 0.64 | 2.79 |
| United States | Chicago | Census tract | 791 | 0.51 | 1.94 |
| United States | New York City | Census tract | 2,198 | 0.18 | 0.87 |
| England | Greater London | MSOA | 963 | 1.14 | 4.12 |
| England | West Midlands/Birmingham | MSOA | 352 | 1.89 | 4.59 |
| England | West Yorkshire/Leeds | MSOA | 297 | 3.06 | 24.45 |

**Supplementary Table 11: Pixel-footprint support per polygon.** Valid intersecting pixels exclude nodata. Effective pixels are the sum of the overlap fractions and are expressed as full-pixel equivalents. Values in parentheses show the 5th and 95th percentiles. Boundary-pixel share uses all geometrically intersecting pixel rectangles.

| Country | $n$ | Intersecting pixels, median (P5–P95) | Effective pixels, median (P5–P95) | Median boundary-pixel share |
|---|---|---|---|---|
| United States | 4,088 | 55.0 (24.0–256.0) | 39.3 (13.0–214.0) | 58.3% |
| England | 1,612 | 313.5 (116.0–1,534.5) | 248.5 (82.7–1,372.5) | 39.4% |

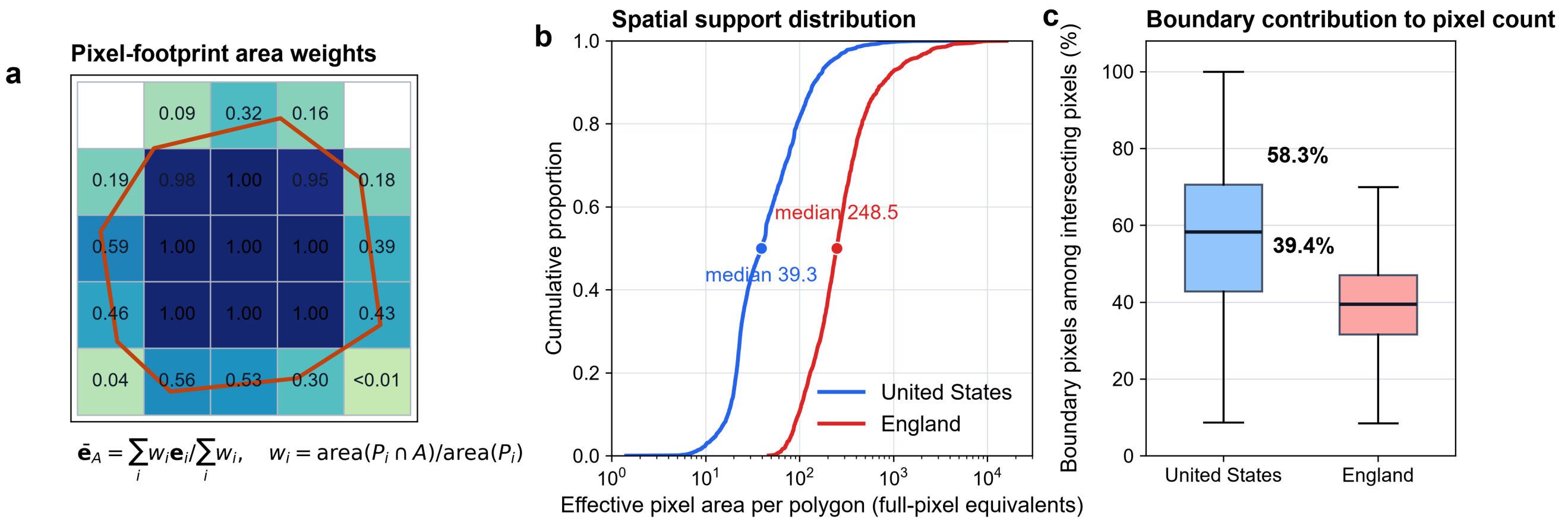

**Supplementary Figure 4: Area-weighted representation construction and observed pixel support.** **a**, Each raster cell is treated as its georeferenced rectangular footprint. Cells located entirely within a polygon receive a weight of 1, whereas boundary cells receive a weight equal to the fraction of their footprint overlapping the polygon. **b**, Empirical cumulative distributions of effective full-pixel equivalents per polygon. **c**, Fraction of intersecting raster cells located on polygon boundaries.

### 3.2.3 Evaluation design

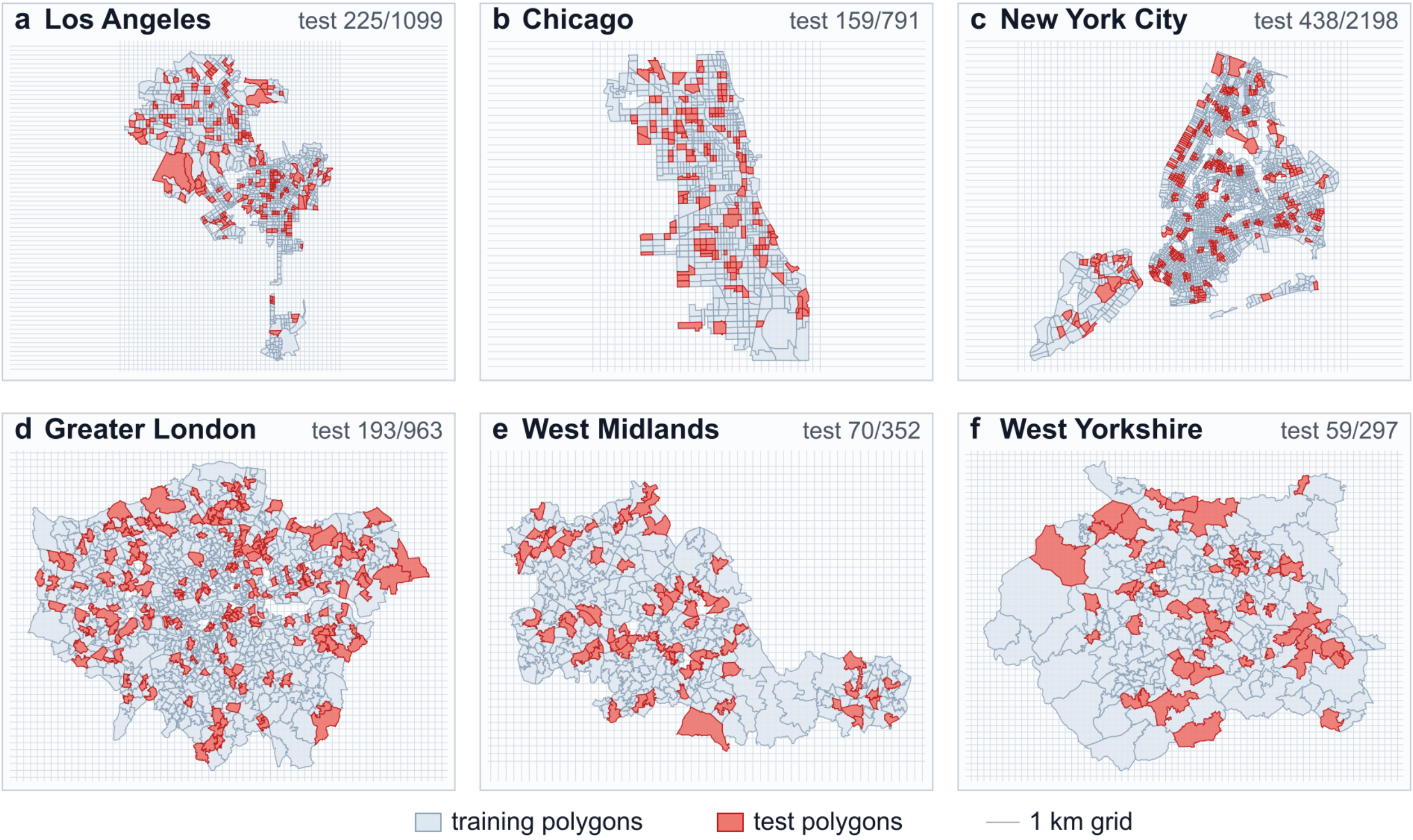

**Supplementary Figure 5: Study regions and an example spatial split.** Blue-grey polygons indicate areas assigned to training, and coral polygons indicate areas used for testing for split seed 0. Thin grey lines show the 1 km grid used to group neighbouring polygon representative coordinates. The same spatial partitioning procedure was repeated using four additional split seeds.

**Spatial split protocol.** We adopted a spatial split strategy for constructing the test set. Specifically, representative coordinates for each polygon were assigned to 1 km grid cells separately within each study region. Grid cells were shuffled reproducibly using five split seeds (seeds 0–4) and assigned to the test set until approximately 20% of the observations in each region had been held out. Complete grid cells, rather than individual polygons, were assigned to one side of the split. 15% of each region's remaining training observations were used for validation.

Identical training and test assignments were used for all three compared methods (coordinate, AEF and PURE). This design evaluates local spatial interpolation while reducing direct train–test mixing among polygons assigned to the same 1 km grid cell. An example partition for split seed 0 is shown in Supplementary Fig. 5.

**Regression model and training.** All socioeconomic indicators were predicted using a multilayer perceptron with four hidden layers containing 512, 256, 128 and 64 units, respectively. Each hidden layer used a rectified linear unit activation, and the final layer contained a single linear output. The same architecture was used for all feature sets. Owing to their different input dimensionalities, the resulting models contained 174,081 trainable parameters for geographic coordinates, 205,825 for AEF and 238,593 for PURE.

Input features and target variables were standardised using statistics calculated from the model-fitting subset only. Input dimensions with a model-fitting-fold standard deviation below $10^{-4}$ were assigned a scale factor of 1, preventing numerical amplification of effectively constant features.

Models were optimised using Adam with a learning rate of 0.002, a batch size of 256 and an L2 regularisation coefficient of $1 \times 10^{-2}$. Training ran for a maximum of 50 epochs, with early stopping after 5 consecutive epochs without improvement in validation RMSE. The best validation checkpoint was restored.

**Evaluation metrics.** Performance was evaluated on the held-out polygons using the coefficient of determination ($R^2$) and mean absolute error (MAE). The main comparison uses $R^2$ because it is scale-free and can therefore be compared across socioeconomic indicators with different units and value ranges.

For each indicator and representation, results were summarised as the mean and standard deviation across the five spatial split seeds. For country-level results, performance was first averaged across the five indicators within each seed; means and standard deviations were then calculated across the five seeds.

### 3.2.4 Results and interpretation

**Overall predictive performance.** Supplementary Fig. 6a shows the $R^2$ performance of the three compared representations, aggregated across all five socioeconomic indicators. PURE achieved higher country-level mean predictive performance than AEF in both national settings. In the United States, PURE achieved a mean $R^2$ of $0.472$, compared with $0.446$ for AEF and $0.004$ for geographic coordinates. The corresponding values in England were $0.548$, $0.462$ and $0.272$, respectively. The mean paired PURE–AEF difference was $+0.026$ in the United States and $+0.086$ in England.

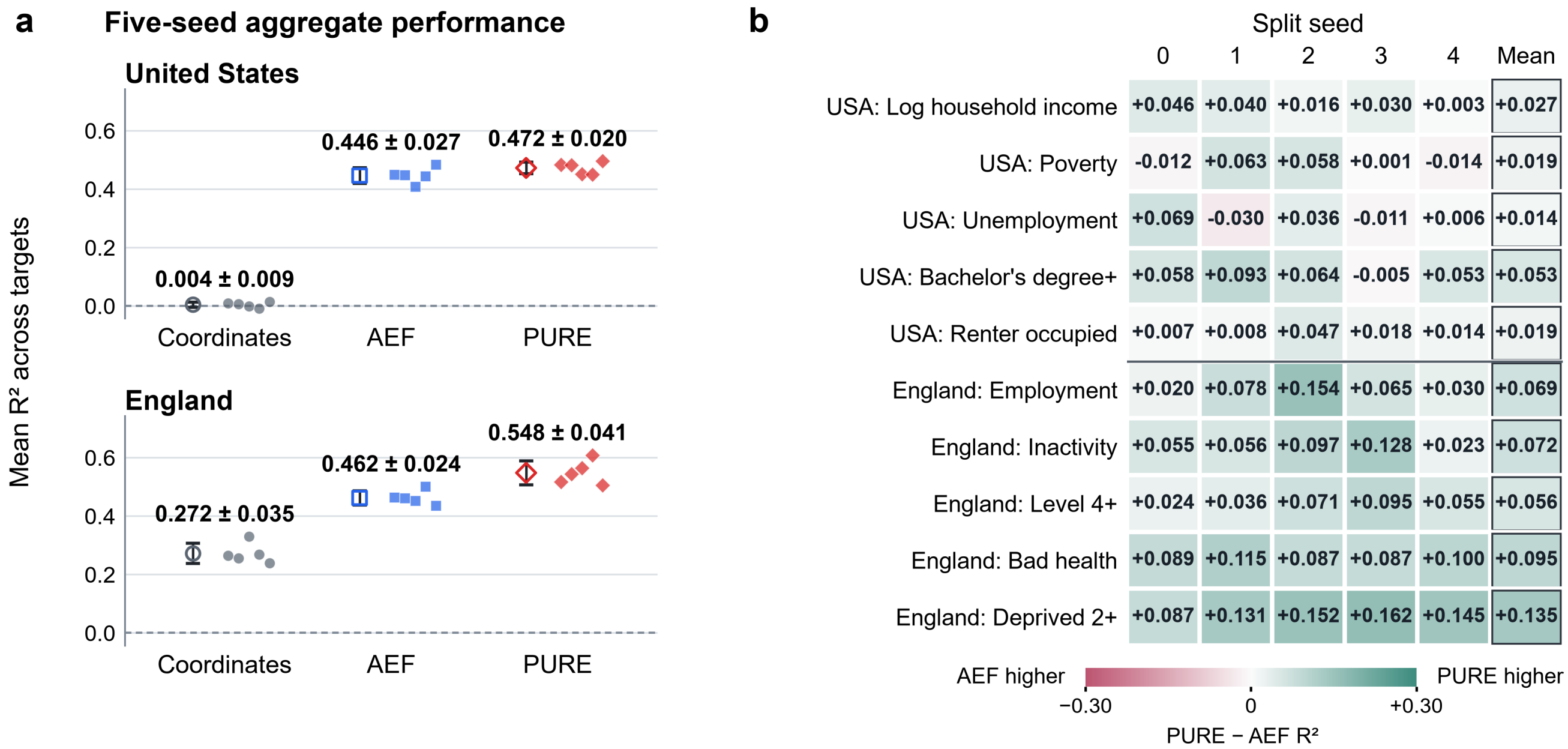

**Supplementary Figure 6: Socioeconomic prediction performance.** **a**, Mean $R^2$ across the five socioeconomic indicators in each country. Small symbols show results for individual spatial split seeds; open symbols to the left indicate means, and whiskers denote standard deviations. **b**, Indicator- and seed-specific differences in $R^2$ between PURE and AEF. Positive values favour PURE.

**Indicator-specific performance.** We next examine indicator-specific performance to provide a more detailed view of how the gains are distributed across individual socioeconomic indicators. As shown in Supplementary Table 12, PURE achieved higher five-seed mean $R^2$ than AEF for all five US indicators and all five English indicators. Supplementary Fig. 6b further shows that, across the 50 matched indicator-by-seed comparisons, PURE achieved higher predictive accuracy in 45 cases (90%).

In the United States, PURE's largest gain was for educational attainment ($\Delta R^2 = +0.053$), followed by log median household income ($+0.027$). Smaller gains were observed for poverty and renter occupancy (both $+0.019$) and unemployment ($+0.014$).

In England, PURE achieved higher mean performance for every indicator. The largest gains were for households deprived in two or more dimensions ($\Delta R^2 = +0.135$), bad or very bad health ($+0.095$) and economic inactivity ($+0.072$). Smaller positive differences were observed for employment ($+0.069$) and Level 4 qualifications or higher ($+0.056$).

Overall, PURE achieved higher mean predictive performance than AEF in both countries and a higher five-seed mean for every socioeconomic indicator. This provides evidence that the benefits of functional enrichment extend beyond the global raster benchmark to official small-area socioeconomic attributes across different national settings.

**Supplementary Table 12: Indicator-specific socioeconomic prediction performance.** Values are mean ± s.d. across five spatial split seeds. Mean absolute error (MAE) is reported in natural-log units for household income and percentage points for all other indicators. Bold values indicate better performance within each indicator: higher mean $R^2$ or lower mean MAE.

| Country | Indicator | AEF $R^2$ | PURE $R^2$ | AEF MAE | PURE MAE |
|---|---|---|---|---|---|
| United States | Log median household income | $0.506 \pm 0.027$ | $\mathbf{0.533 \pm 0.023}$ | $0.255 \pm 0.006$ | $\mathbf{0.249 \pm 0.003}$ |
| United States | Poverty rate | $0.302 \pm 0.060$ | $\mathbf{0.321 \pm 0.050}$ | $7.233 \pm 0.258$ | $\mathbf{7.035 \pm 0.152}$ |
| United States | Unemployment rate | $0.143 \pm 0.036$ | $\mathbf{0.157 \pm 0.049}$ | $3.808 \pm 0.090$ | $\mathbf{3.763 \pm 0.083}$ |
| United States | Bachelor's degree or higher | $0.638 \pm 0.027$ | $\mathbf{0.690 \pm 0.033}$ | $10.563 \pm 0.241$ | $\mathbf{9.648 \pm 0.538}$ |
| United States | Renter-occupied households | $0.640 \pm 0.025$ | $\mathbf{0.658 \pm 0.021}$ | $11.747 \pm 0.145$ | $\mathbf{11.381 \pm 0.272}$ |
| England | Employment rate | $0.413 \pm 0.061$ | $\mathbf{0.482 \pm 0.061}$ | $3.799 \pm 0.201$ | $\mathbf{3.504 \pm 0.058}$ |
| England | Economic inactivity rate | $0.452 \pm 0.037$ | $\mathbf{0.524 \pm 0.070}$ | $3.436 \pm 0.174$ | $\mathbf{3.118 \pm 0.181}$ |
| England | Level 4 qualification or higher | $0.731 \pm 0.018$ | $\mathbf{0.787 \pm 0.016}$ | $5.576 \pm 0.396$ | $\mathbf{4.914 \pm 0.397}$ |
| England | Bad or very bad health | $0.357 \pm 0.042$ | $\mathbf{0.453 \pm 0.038}$ | $0.955 \pm 0.032$ | $\mathbf{0.862 \pm 0.027}$ |
| England | Households deprived in two or more dimensions | $0.358 \pm 0.027$ | $\mathbf{0.493 \pm 0.045}$ | $4.622 \pm 0.145$ | $\mathbf{3.974 \pm 0.197}$ |

## 3.3 Commuting flow generation

### 3.3.1 Motivation and prediction task

Mobility flow generation estimates interactions between origins and destinations from their geographic and contextual characteristics[10,11]. We further evaluated the urban representations using a supervised commuting flow generation task at the US Census tract level. In contrast to the socioeconomic regression task, which predicts an attribute of an individual spatial unit, this task models directed interactions between residential origins and workplace destinations.

For each origin tract $o$, the model received its embedding, the embeddings of a set of candidate destination tracts $d \in \mathcal{D}_o$, and the corresponding origin–destination distances. The model produced a probability distribution over the candidate destinations. During evaluation, predicted probabilities were converted to flow counts by multiplying them by the observed total outflow from the origin:

$$\hat{y}_{od} = T_o p_{od}, \qquad T_o = \sum_{d \in \mathcal{D}_o} y_{od}, \tag{1}$$

where $p_{od}$ is the predicted probability of travelling from origin $o$ to destination $d$, $y_{od}$ is the observed flow and $\hat{y}_{od}$ is the predicted flow.

### 3.3.2 Data and preprocessing

We constructed a Census tract-level commuting flow benchmark covering five large Metropolitan Statistical Areas (MSAs) in the United States. The study regions were selected from among the ten largest US MSAs and comprised New York–Newark–Jersey City, Los Angeles–Long Beach–Anaheim, Chicago–Naperville–Elgin, Dallas–Fort Worth–Arlington and Washington–Arlington–Alexandria. Together, these regions provide variation in metropolitan population, spatial extent and urban structure while retaining consistent national sources of commuting flow and boundary data.

For brevity, the five MSAs are referred to throughout the Supplementary Information by the shorthand names: New York, Los Angeles, Chicago, Dallas and Washington. Their full names, geographic coverage and final numbers of retained Census tracts are provided in Supplementary Table 13.

**Supplementary Table 13: Study regions included in the US commuting flow benchmark.** Only Census tracts with finite representations available from both AEF and PURE were retained. The shorthand names are used throughout the Supplementary Information.

| Metropolitan Statistical Area | Shorthand | State coverage | Retained Census tracts |
|---|---|---|---|
| New York–Newark–Jersey City | New York | NY–NJ–PA | 5,388 |
| Los Angeles–Long Beach–Anaheim | Los Angeles | CA | 3,109 |
| Chicago–Naperville–Elgin | Chicago | IL–IN–WI | 2,335 |
| Dallas–Fort Worth–Arlington | Dallas | TX | 1,721 |
| Washington–Arlington–Alexandria | Washington | DC–VA–MD–WV | 1,483 |

**Commuting flow and boundary data.** The commuting flow data were obtained from the 2023 LEHD Origin–Destination Employment Statistics dataset[12]. The original records describe commuting flows between residential and workplace Census blocks. These block-to-block flows were aggregated to Census tracts by taking the first 11 digits of `h_geocode` as the residential origin tract and the first 11 digits of `w_geocode` as the workplace destination tract. The `S000` field, which records the total number of jobs, was used as the flow count.

The resulting data define a directed, weighted origin–destination network within each MSA. Each origin tract may be linked to multiple workplace destination tracts, with the weight of each connection given by the observed number of commuters.

Census tract geometries were obtained from the 2023 TIGER/Line tract boundary files[13]. Tracts were assigned to each MSA using county-level metropolitan definitions. Only tracts with finite AEF and PURE representations were retained, ensuring that the two representations were evaluated using an identical set of residential origins and workplace destinations.

**Tract representations.** The spatial unit of this task is the US Census tract, which is also used as the spatial unit in the small-area socioeconomic prediction task in Section 3.2. As in that task, we use the same pretrained AEF and PURE representations and apply the same pixel-footprint area-weighted averaging procedure described in Section 3.2.2 to derive the representation for each Census tract in the 5 selected MSAs.

#### 3.3.3 Evaluation design

**Data partitioning.** For each MSA, five random tract-level partitions were generated. Census tracts were divided into training, validation and test sets using a 65%/15%/20% ratio. Origin–destination records were then retained only when both the origin and destination tracts belonged to the same partition. This prevented flows involving tracts assigned to different partitions from being used during model fitting or evaluation.

Training and validation data samples included observed positive destinations together with randomly sampled zero-flow destinations. The size of the sampled candidate set was determined from the 99.5th percentile of the observed number of destinations across origin tracts. Evaluation used the complete destination set. For every test origin with positive observed total outflow, all tracts in the test partition were included as candidate destinations, including those with zero observed flow.

**Distance feature.** Great-circle distances were calculated between the centroids of the origin and destination tracts. Within each MSA, distances were divided by the maximum pairwise tract distance so that the resulting auxiliary feature was normalised to a common range.

**Flow generation models.** AEF and PURE embeddings were evaluated using two commonly used downstream models: DeepGravity [10] and a bilinear interaction model, denoted as DG and Bilinear, respectively, in the subsequent table and figure presenting the evaluation results. The former provides a flexible nonlinear model of origin, destination and distance features, whereas the latter provides a more constrained formulation of interactions between origin and destination representations and has been used as a common downstream model in previous studies [14,15]. Both models generated scores for the candidate destinations, which were normalised to form a destination probability distribution for each origin. Models were trained separately for each MSA, representation and tract-level data split. One model-initialisation seed was used for each split. At evaluation, destination scores were normalised across the complete set of test-partition destinations for each eligible origin, and performance was calculated over the resulting full origin–destination candidate set.

**Evaluation metrics.** Flow generation performance was measured using 5 common metrics: common part of commuters (CPC), MAE, RMSE, Pearson correlation and Jensen–Shannon distance (JSD). In particular, CPC measures the overlap between observed and predicted flows:

$$\mathrm{CPC} = \frac{2\sum_{(o,d)} \min\left(y_{od}, \hat{y}_{od}\right)}{\sum_{(o,d)} y_{od} + \sum_{(o,d)} \hat{y}_{od}}. \tag{2}$$

MAE and RMSE measure errors in predicted flow counts, while Pearson correlation measures agreement in the relative variation of observed and predicted flows. JSD measures the discrepancy between their normalised flow distributions. Higher CPC and Pearson correlation indicate better performance, whereas lower MAE, RMSE and JSD indicate better performance.

Results were summarised as the mean and standard deviation across the five tract-level data splits for each metropolitan area, representation and downstream model architecture.

#### 3.3.4 Results and analyses

Supplementary Table 14 presents the overall commuting flow generation results across the five metropolitan areas. From the table, we can see that PURE consistently outperforms AEF across different study areas, downstream models and evaluation metrics. For all ten metropolitan area–model combinations, PURE achieves higher mean CPC and Pearson correlation, while also producing lower MAE, RMSE and JSD. Although the magnitude of improvement varies across metropolitan areas, the direction of the change is remarkably consistent. This suggests that the functional information incorporated by PURE provides a consistent advantage for commuting flow generation across different study areas and downstream models.

**Supplementary Table 14: Commuting flow generation performance over the full candidate destination set.** Values are mean ± s.d. across five random tract-level train–validation–test splits, using one fixed model seed per split. For every eligible test origin, predictions are normalised across all test-split destinations, including OD pairs with zero observed flow. AEF and PURE use identical tract cohorts and splits within each metropolitan area. Higher CPC and Pearson $r$ are better; lower MAE, RMSE and JSD are better. Bold values identify the better representation within each metropolitan area and downstream model. Rep. denotes representation.

| Metropolitan Area | Model | Rep. | CPC ↑ | MAE ↓ | RMSE ↓ | Pearson $r$ ↑ | JSD ↓ |
|---|---|---|---|---|---|---|---|
| New York | DG | AEF | $0.301 \pm 0.020$ | $0.466 \pm 0.023$ | $2.462 \pm 0.115$ | $0.634 \pm 0.013$ | $0.605 \pm 0.013$ |
| | | PURE | $\mathbf{0.310 \pm 0.028}$ | $\mathbf{0.459 \pm 0.019}$ | $\mathbf{2.423 \pm 0.111}$ | $\mathbf{0.659 \pm 0.020}$ | $\mathbf{0.599 \pm 0.018}$ |
| | Bilinear | AEF | $0.250 \pm 0.020$ | $0.504 \pm 0.023$ | $2.664 \pm 0.163$ | $0.503 \pm 0.031$ | $0.641 \pm 0.013$ |
| | | PURE | $\mathbf{0.268 \pm 0.013}$ | $\mathbf{0.491 \pm 0.025}$ | $\mathbf{2.595 \pm 0.140}$ | $\mathbf{0.551 \pm 0.037}$ | $\mathbf{0.627 \pm 0.009}$ |
| Los Angeles | DG | AEF | $0.326 \pm 0.006$ | $0.830 \pm 0.100$ | $3.126 \pm 0.476$ | $0.638 \pm 0.022$ | $0.590 \pm 0.003$ |
| | | PURE | $\mathbf{0.351 \pm 0.017}$ | $\mathbf{0.796 \pm 0.082}$ | $\mathbf{2.950 \pm 0.415}$ | $\mathbf{0.707 \pm 0.036}$ | $\mathbf{0.573 \pm 0.010}$ |
| | Bilinear | AEF | $0.289 \pm 0.011$ | $0.877 \pm 0.101$ | $3.304 \pm 0.482$ | $0.496 \pm 0.108$ | $0.618 \pm 0.008$ |
| | | PURE | $\mathbf{0.315 \pm 0.016}$ | $\mathbf{0.845 \pm 0.094}$ | $\mathbf{3.119 \pm 0.466}$ | $\mathbf{0.612 \pm 0.047}$ | $\mathbf{0.598 \pm 0.010}$ |
| Chicago | DG | AEF | $0.340 \pm 0.025$ | $0.929 \pm 0.098$ | $3.172 \pm 0.261$ | $0.545 \pm 0.029$ | $0.580 \pm 0.017$ |
| | | PURE | $\mathbf{0.368 \pm 0.028}$ | $\mathbf{0.883 \pm 0.090}$ | $\mathbf{3.006 \pm 0.269}$ | $\mathbf{0.618 \pm 0.029}$ | $\mathbf{0.561 \pm 0.019}$ |
| | Bilinear | AEF | $0.274 \pm 0.021$ | $1.041 \pm 0.123$ | $3.489 \pm 0.317$ | $0.370 \pm 0.071$ | $0.628 \pm 0.014$ |
| | | PURE | $\mathbf{0.308 \pm 0.024}$ | $\mathbf{0.989 \pm 0.114}$ | $\mathbf{3.356 \pm 0.290}$ | $\mathbf{0.465 \pm 0.048}$ | $\mathbf{0.603 \pm 0.018}$ |
| Dallas | DG | AEF | $0.430 \pm 0.012$ | $1.409 \pm 0.251$ | $4.324 \pm 0.717$ | $0.613 \pm 0.043$ | $0.519 \pm 0.008$ |
| | | PURE | $\mathbf{0.464 \pm 0.028}$ | $\mathbf{1.312 \pm 0.181}$ | $\mathbf{4.088 \pm 0.587}$ | $\mathbf{0.636 \pm 0.042}$ | $\mathbf{0.495 \pm 0.021}$ |
| | Bilinear | AEF | $0.378 \pm 0.026$ | $1.539 \pm 0.222$ | $4.724 \pm 0.678$ | $0.468 \pm 0.056$ | $0.559 \pm 0.020$ |
| | | PURE | $\mathbf{0.400 \pm 0.025}$ | $\mathbf{1.485 \pm 0.217}$ | $\mathbf{4.601 \pm 0.698}$ | $\mathbf{0.496 \pm 0.081}$ | $\mathbf{0.541 \pm 0.018}$ |
| Washington | DG | AEF | $0.310 \pm 0.015$ | $1.401 \pm 0.166$ | $4.712 \pm 0.706$ | $0.538 \pm 0.021$ | $0.599 \pm 0.010$ |
| | | PURE | $\mathbf{0.356 \pm 0.029}$ | $\mathbf{1.306 \pm 0.145}$ | $\mathbf{4.451 \pm 0.595}$ | $\mathbf{0.582 \pm 0.031}$ | $\mathbf{0.569 \pm 0.019}$ |
| | Bilinear | AEF | $0.239 \pm 0.010$ | $1.576 \pm 0.142$ | $5.112 \pm 0.614$ | $0.361 \pm 0.023$ | $0.649 \pm 0.006$ |
| | | PURE | $\mathbf{0.295 \pm 0.015}$ | $\mathbf{1.461 \pm 0.144}$ | $\mathbf{4.909 \pm 0.648}$ | $\mathbf{0.437 \pm 0.044}$ | $\mathbf{0.612 \pm 0.010}$ |

The two downstream models differ in their overall predictive performance, with DeepGravity generally achieving stronger absolute results than the simpler bilinear model. However, this difference is not the main focus of the comparison. More importantly, the advantage of PURE over AEF is observed under both model architectures and across all five metropolitan areas. We argue that this consistency provides evidence that the improvement is associated with the urban representations themselves, rather than being dependent on a particular downstream flow-generation model.

Supplementary Fig. 7 provides a closer look at these results by showing the CPC differences between PURE and AEF for each individual spatial split. The figure shows that PURE achieves higher CPC in 47 of the 50 paired comparisons, corresponding to 94% of all metropolitan area–model–split combinations. Positive differences are observed across all five metropolitan areas and under both downstream models, while the three negative cases are comparatively small. The mean CPC difference is also positive for every metropolitan area under both architectures. This finer-grained comparison suggests that the improvements reported in Supplementary Table 14 are not driven by a small number of favourable data partitions, but remain broadly consistent across different spatial splits.

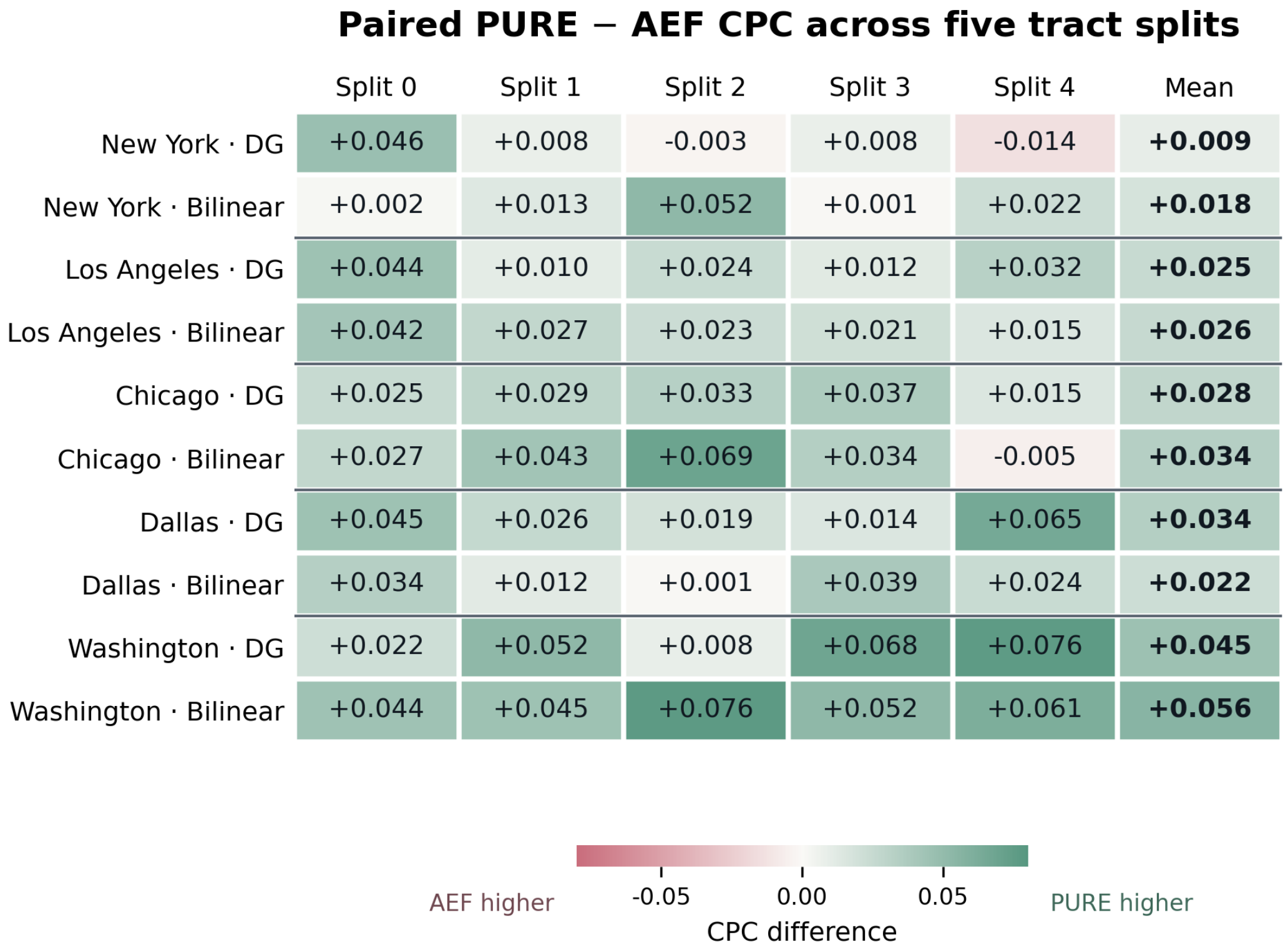

**Supplementary Figure 7: Paired CPC differences between PURE and AEF across tract-level data splits.** Each cell shows the difference in CPC between PURE and AEF for one of the five tract-level splits within each metropolitan area and downstream architecture. Positive values indicate higher CPC for PURE. The final column reports the mean difference across the five splits. DG denotes DeepGravity.

### 3.4 Summary across downstream tasks

The two independent downstream evaluations examine different levels of urban information. The socioeconomic task tests whether a representation supports prediction of attributes associated with individual neighbourhoods, whereas the commuting flow task tests whether it supports modelling of directed interactions between residential and workplace locations.

PURE achieved higher country-level mean socioeconomic prediction performance than AEF in both national settings, although the magnitude and consistency of the gains varied across indicators. In the commuting flow task, PURE improved performance across all five US metropolitan areas under two commonly used downstream model architectures. Together, these results provide complementary evidence that the functional semantics introduced by PURE are useful for both small-area attribute prediction and the relatively more complex human mobility modelling.